\documentclass[journal]{IEEEtran}
\usepackage{longfigure}
\usepackage{flushend}
\usepackage{forest}
\usepackage{diagbox}
\usepackage{slashbox}
\usepackage{threeparttable}
\usepackage{fontawesome}
\usepackage{multirow}
\usepackage{multicol}
\usepackage{cuted}
\usepackage{fullpage}
\usepackage{bbm}
\usepackage{enumitem}
\usepackage{caption} 
\usepackage{balance} 
\usepackage{hyperref}
\usepackage{amsthm}
\usepackage[square, comma, sort&compress, numbers]{natbib}
\usepackage{amsmath,amssymb,amsfonts}
\usepackage{algorithmic}
\usepackage{color, colortbl}
\usepackage{array, multirow, graphicx}
\usepackage{textcomp}
\usepackage{subfigure}
\usepackage{float}
\usepackage{xcolor}

\usepackage{graphicx}
\usepackage{pifont} 
\usepackage{boxhandler}
\usepackage[export]{adjustbox}

\usepackage[T1]{fontenc}
\usepackage[utf8]{inputenc}
\usepackage{tikz}

\newcommand{\specialcell}[2][l]{%
  \begin{tabular}[#1]{@{}l@{}}#2\end{tabular}}
  
\definecolor{greenish}{rgb}{0.26, 0.586, 0.16}
\definecolor{reddish}{rgb}{0.59, 0, 0.09} 
\makeatletter
\ifCLASSINFOpdf
\else
\fi
\begin{document}
%
\title{Robust Clustering using Hyperdimensional Computing}

%
%
%

\author{Lulu~Ge,~\IEEEmembership{Student Member,~IEEE,}
        and Keshab K. Parhi,~\IEEEmembership{Fellow,~IEEE}
\thanks{L. Ge and K.K. Parhi are with the Department of Electrical and Computer Engineering, University of Minnesota,
Minneapolis, MN 55455, USA e-mail: \{ge000567, parhi\}@umn.edu}
\thanks{This work was supported in parts by CISCO Systems, and by the Chinese Scholarship Council (CSC).}}

%
%

\markboth{Journal of \LaTeX\ Class Files,~Vol.~XX, No.~XX, Month~Year}%
{Shell \MakeLowercase{\textit{et al.}}: Bare Demo of IEEEtran.cls for IEEE Journals}
%



\maketitle

\begin{abstract}
This paper addresses the clustering of data in the hyperdimensional computing (HDC) domain. In prior work, an HDC-based clustering framework, referred to as HDCluster, has been proposed. However, the performance of the existing HDCluster is {\em not} robust. The performance of HDCluster is degraded as the hypervectors for the clusters are chosen at random during the initialization step. To overcome this bottleneck, we assign the initial cluster hypervectors by exploring the similarity of the encoded data, referred to as \textit{query} hypervectors. Intra-cluster hypervectors have a higher similarity than inter-cluster hypervectors. Harnessing the similarity results among query hypervectors, this paper proposes four HDC-based clustering algorithms: similarity-based k-means, equal bin-width histogram, equal bin-height histogram, and similarity-based affinity propagation. 
Experimental results illustrate that: (i) Compared to the existing HDCluster, our proposed HDC-based clustering algorithms can achieve better accuracy, more robust performance, fewer iterations, and less execution time. Similarity-based affinity propagation outperforms the other three HDC-based clustering algorithms on eight datasets by $ 2\%\sim38\%$ in clustering accuracy. 
(ii) Even for one-pass clustering, i.e., without any iterative update of the cluster hypervectors, our proposed algorithms can provide more robust clustering accuracy than HDCluster.
(iii) Over eight datasets, five out of eight can achieve higher or comparable accuracy when projected onto the hyperdimensional space. Traditional clustering is more desirable than HDC when the number of clusters, $k$, is large.
\end{abstract} 


%
\IEEEpeerreviewmaketitle

\section{Introduction}\label{sec:intro}


Hyperdimensional computing (HDC) is a novel computing paradigm that mimics brain behavior \cite{kanerva1988sparse, kanerva1997fully, kanerva2009hyperdimensional}. In general, HDC employs its unique data type in the hyperdimensional space -- \textit{hypervectors}, which are ultra-long vectors and usually have a dimensionality $d$ of a thousand bits, e.g., $d = 10,000$. Many data structures, such as  letters \cite{joshi2016language, rahimi2016robust, imani2018hdna}, signals \cite{burrello2019laelaps, hersche2018exploring, imani2017voicehd, ge2021seizure, ge2022applicability}, graphs \cite{poduval2022graphd, kleyko2023survey,kang2022relhd}, and images \cite{manabat2019performance,kleyko2016holographic,plate1995holographic} can be represented using HDC. Current research findings have demonstrated that HDC can achieve comparable performance with traditional machine learning techniques but support few-shot learning \cite{rahimi2017high, rahimi2017hyperdimensional,imani2018hierarchical, rahimi2018hyperdimensional,burrello2018one,rahimi2018efficient}, high energy efficiency \cite{imani2017low,gupta2018felix,imani2019adapthd, imani2019bric,morris2019comphd,imani2019sparsehd,hersche2020integrating,karunaratne2021energy,guo2021hyperrec,basaklar2021hypervector,zhao2022fedhd}, and hardware acceleration \cite{montagna2018pulp, schmuck2019hardware,kang2022xcelhd}. HDC has wide applications that are not limited to supervised learning (e.g., classification \cite{ge2020classification,ung2022premature, billmeyer2021biological} and regression \cite{hernandez2021reghd}), unsupervised learning (e.g., clustering \cite{kaski1998dimensionality, imani2019hdcluster, gupta2022store,morris2022HyDREA}), and even reasoning \cite{mitrokhin2020symbolic, widdows2015reasoning, kleyko2018classification}.


\begin{table*}[ht]
  \centering
  \caption{Our replication results of HDCluster as compared to HDCluster's published performance in \cite{imani2019hdcluster}.}
  \setlength\tabcolsep{0.5pt}
  \begin{adjustbox}{angle = 0, width=1\textwidth}
    \begin{tabular}{lccccccccc}
    \rowcolor{black} \textcolor{white}{\textbf{Datasets}} & \textcolor{white}{\href{https://doi.org/10.24432/C53K8Q}{\textbf{MNIST}}} & \textcolor{white}{\href{https://doi.org/10.24432/C51G69}{\textbf{ISOLET}}} & \textcolor{white}{\href{https://doi.org/10.24432/C56C76}{\textbf{IRIS}}} & \textcolor{white}{\href{https://doi.org/10.24432/C5WW2P}{\textbf{Glass}}} & \textcolor{white}{\textbf{Unbalance}} & \textcolor{white}{\href{https://doi.org/10.24432/C5R88H}{\textbf{RNA-seq}}} & \textcolor{white}{\href{https://doi.org/10.24432/C5DW2B}{\textbf{Cancer}}} & \textcolor{white}{\href{https://doi.org/10.24432/C5388M}{\textbf{Ecoli}}} & \textcolor{white}{\href{https://doi.org/10.24432/C59C74}{\textbf{Parkinsons}}} \\
    \rowcolor{black}  & \textcolor{white}{[\%]}& \textcolor{white}{[\%]}& \textcolor{white}{[\%]}& \textcolor{white}{[\%]}&\textcolor{white}{[\%]}& \textcolor{white}{[\%]}& \textcolor{white}{[\%]}& \textcolor{white}{[\%]}& \textcolor{white}{[\%]}\\
    k-means \cite{imani2019hdcluster} & 48.8 & 28.4 & 88.7 & 51.8 & 93.8 & 37.5 & 94.1 & 74.3  & 75.3  \\
    HDCluster \cite{imani2019hdcluster} & 58.6 & 33.1 & 89.9 & 67.5 & 92.3 & 37.5 & 96.2 & 78.5  & 75.6  \\
    \noalign{\hrule height 1pt}
     Replication* &  47.10($\pm$8.79) & 32.26($\pm$8.11) & 72.39($\pm$13.85) & 43.17($\pm$6.07) & /     & 59.01($\pm$16.05) & 73.50($\pm$10.44) & 53.49($\pm$10.80) & 75.45($\pm$0.47) \\
    \noalign{\hrule height 1pt}
   \multicolumn{10}{l}{\specialcell[b]{$^*$: displays our replicated clustering accuracy [\%] of HDCluter over 500 runs in the format of $[$mean ($\pm$ standard deviation)$]$. \\Symbol ``/'' indicates that the ``Unbalance'' dataset is not publicly available. Therefore, there are no replicated results.}}
    \end{tabular}\label{t:research_gap}
    \end{adjustbox}
\end{table*}

Clustering is one of the fundamental tasks in machine learning that seeks to create clusters/groups of similar data. 
For clustering using HDC, \cite{kaski1998dimensionality} explains analytically and empirically why the random mapping in HDC data representation can approximately preserve the mutual similarity among the original data. As an alternative to dimensionality reduction, random projection has been demonstrated to achieve comparable data separability with negligible computational complexity with principal component analysis (PCA) for document classification. Therefore, as indicated by \cite{kaski1998dimensionality}, HDC is promising for faster clustering, especially in situations when the original data has a huge dimensionality. An approach, referred to as HDCluster, was presented in \cite{imani2019hdcluster} to cluster data in the HDC domain by mimicking the traditional k-means. Furthermore, \cite{imani2019hdcluster} showed that HDCluster can be more accurate than traditional k-means over diverse datasets (see Table \ref{t:research_gap}). An in-storage computing solution, called Store-n-Learn, is proposed for HDC-based classification and clustering across the flash hierarchy in \cite{gupta2022store}. Experiments show that Store-n-Learn can achieve on average $543\times$ faster than CPU for clustering over ten datasets. In \cite{morris2022HyDREA}, a processing-in-memory (PIM) architecture that utilizes HDC for a more robust and efficient machine learning system was proposed. In particular, clustering is supported by HyDREA to achieve $32\times$ speed up and $289\times$ energy efficiency than the baseline architecture. It may be noted that the HDC-based clustering in \cite{kaski1998dimensionality,imani2019hdcluster} is algorithm-oriented, whereas \cite{gupta2022store,morris2022HyDREA} emphasize the hardware implementations using the same clustering framework as \cite{imani2019hdcluster}.

\begin{figure}
    \centering
    \includegraphics[width=\linewidth]{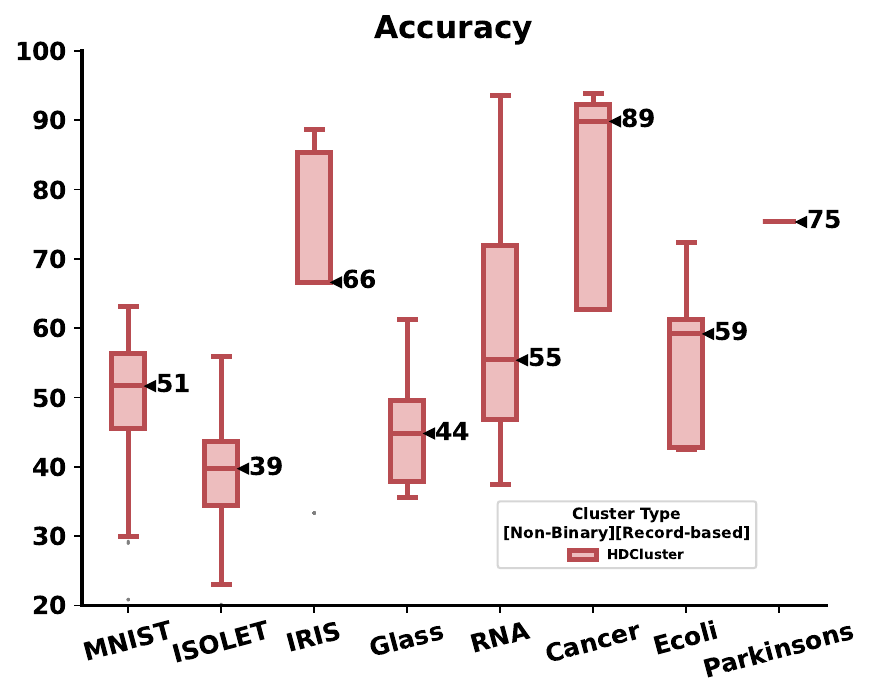}
    \captionsetup{justification=justified,singlelinecheck=false}
    \caption{Replicated clustering performance using HDCluster over 500 runs using versatile machine learning datasets in \cite{imani2019hdcluster}. Different runs employ various random seeds to generate the random hypervectors serving as the initial $k$ cluster hypervectors.}\label{fig:hdcluster_boxplot}
\end{figure}

The performance of the existing HDCluster is {\em not} robust. The performance of HDCluster is degraded as the hypervectors for the clusters are chosen at random during the initialization step. In this approach, the variance of the clustering performance is dependent on the selection of the initial hypervector.
To be more specific, for a $k$-cluster problem, the possible selection for $k$ binary/bipolar random hypervectors from the hyperdimensional space is $2^d \choose k$, where $d$ is the dimensionality of the hypervectors. In HDCluster, the initialization of $k$ cluster hypervectors plays a significant role in the quality of clustering performance, i.e., accuracy and number of iterations. 
Table \ref{t:research_gap} lists the clustering accuracy (summarized in the [mean$\pm$std] manner) of HDCluster and traditional $k$-means in \cite{imani2019hdcluster}. Our replication results for HDCluster over 500 runs using different random seeds show that the variance for HDCluster accuracy is high, which has not been reported in \cite{imani2019hdcluster}. Such a high variance in clustering performance is vividly displayed in Fig. \ref{fig:hdcluster_boxplot}. Furthermore, the number of iterations required for convergence in HDCluster always exceeds 300. This necessitates the design of new clustering algorithms in the HDC domain that are robust. In this paper, we propose four novel HDC-based clustering algorithms where the initialization step assigns cluster hypervectors by exploiting the similarity of the data in the encoded domain. These encoded data are referred to as \textit{query} hypervectors.

In clustering, intra-cluster hypervectors are much closer/similar than inter-cluster hypervectors. Therefore, the categorized information can be inferred by the similarity results among these query hypervectors. In this paper, four HDC-based clustering algorithms are proposed: similarity-based k-means, equal bin-width histogram, equal bin-height histogram, and similarity-based affinity propagation. The first three algorithms require only one-dimensional similarity results, while the fourth algorithm requires a matrix of similarity results. In this paper, the emphasis is more on the algorithms than on the hardware implementation.

The contribution of this paper can be summarized as follows:
(i) Using HDCluster, the selection of random hypervectors as cluster hypervectors in hyperdimensional space can cause a high variance in clustering performance which can impact the reliability and consistency of clustering results. Note that this has not been mentioned in the HDCluster paper.
(ii) In contrast to HDCluster's random assignment for the initial cluster hypervectors in the data space (hyperdimensional space), we learn from the data themselves by leveraging the fact that intra-cluster hypervectors have a higher similarity than inter-cluster hypervectors. To achieve this, the similarity results among query hypervectors are utilized. Our proposed HDC-based clustering algorithms are more robust in clustering performance. In addition, our algorithms achieve higher clustering accuracy, fewer iterations for updating cluster hypervectors, and less program execution time as compared to the existing HDCluster. Particularly, similarity-based affinity propagation in the HDC domain always achieves higher accuracy than the traditional k-means, hierarchical clustering, and other HDC-based clustering algorithms over all tested eight datasets.
(iii) The effectiveness of the projection onto hyperdimensional space on clustering performance is examined by applying three traditional clustering algorithms to both the original data and the encoded data. According to the experimental results over eight datasets, the HDC domain-based approaches can achieve similar, if not necessarily better, performance in five out of eight datasets, as compared to the original domain. Additionally, we observe that clustering using the original space is preferable if the number of clusters, $k$, is large.

The remainder of this paper is organized as follows. Section \ref{sec:pre} reviews three traditional clustering algorithms. This section also presents the background of HDC and gives a brief overview of the existing HDCluster. Section \ref{sec:method} illustrates our proposed HDC-based clustering algorithms. The experimental results of our proposed algorithms are compared with the existing HDCluster approach in Sec. \ref{sec:results}. Apart from the update of cluster hypervectors, one-pass clustering is also investigated in this section. A discussion of whether the projection of the original data onto the hyperdimensional space is helpful or not is also provided. Possible future directions of HDC-based clustering are illustrated in Sec. \ref{sec:results}. Finally, Section \ref{sec:con} concludes the paper.

\section{Preliminaries} \label{sec:pre}
In this section, three traditional clustering algorithms are briefly reviewed. Then we introduce the basics of HDC and review the framework of the existing HDCluster. 

\subsection{Traditional Clustering Algorithms}
\subsubsection{\textbf{Traditional k-means}} To create $k$ clusters from the given $N$ data points, as a simple and efficient algorithm, the traditional k-means is widely used to find the local optimal solution \cite{arthur2007k}. Its goal is to minimize the sum of the squared distances (denoted as $\phi$) between every data point and its associated cluster center. In k-means, the initial $k$ clusters are randomly selected from the data domain. \ding{202} Each data point is then assigned to the closest cluster center. \ding{203} After all data points are assigned, each cluster center is recomputed/updated by the mean of its constituent data points. Repeat \ding{202}-\ding{203} until $\phi$ converges or this algorithm exceeds the pre-defined maximum iteration number.

\subsubsection{\textbf{Traditional Hierarchical Clustering}} 
This algorithm aims to build the hierarchy of clusters so that the clusters are organized in a tree-like structure \cite{johnson1967hierarchical, murtagh2012algorithms, murtagh2017algorithms}. There are two typical methods to conduct hierarchical clustering: bottom-up and top-down. In general, the top-down method is more computationally expensive and may not always produce well-defined clusters. Thus in this paper, we use bottom-up hierarchical clustering. Simply stated, bottom-up hierarchical clustering starts with an $N \times N$ matrix of pairwise distances that are computed from the given $N$ data samples. Initially, each data sample is viewed as a unique cluster. Therefore, there are $N$ clusters at the very beginning. \ding{202} Pairs of clusters that have the closest distance are merged as a new cluster so that this new cluster is computed as the average of all the data points that belong to the merged clusters. \ding{203} The size of the matrix of distances is reduced by $1$ and this distance matrix should be recalculated. Repeat \ding{202}-\ding{203} until all the data points are merged into one cluster. 

\subsubsection{\textbf{Traditional Affinity Propagation}} This algorithm identifies a subset of representative examples (called ``\textit{exemplars}'') from the clustered data by passing messages between the data points. It takes the $N \times N$ similarity matrix between all data points as an input. Two kinds of messages are exchanged between data points: \textit{responsibility} and \textit{availability}. The responsibility ($r(i, k)$) reflects how well-suited the point $k$ is to serve as the exemplar for point $i$, whereas the availability ($a(i, k)$) represents how appropriate the point $i$ chooses the point $k$ as its exemplar. Both responsibility and availability are iteratively updated based on the messages passed between data points until this algorithm converges. Finally, each data point is assigned to a cluster based on its exemplar. Interested readers are referred to \cite{frey2007clustering} for more details. 

\subsection{HDC Background}

\subsubsection{\textbf{Hypervectors}}
Using the \textit{seed} hypervectors, also called the  \textit{base}/\textit{basis}/\textit{atomic} hypervectors, HDC maps the original data onto the hyperdimensional space. With an encoding approach, those seed hypervectors are manipulated to form a \textit{compound}/\textit{composite} hypervector, which corresponds to the input data. The basic arithmetic manipulations involved in HDC are nothing but three point-wise operations: addition (+), multiplication (*), and permutation ($\rho$). In addition, a majority rule \cite{ge2020classification} is required to ensure each bit of the hypervectors is binary ($\{0,1\}$) or bipolar ($\{-1,1\}$) for binary/bipolar HDC.

In this paper, the seed hypervectors are either \textit{random} hypervectors or \textit{level} hypervectors: (i) random hypervectors are quasi-orthogonal to each other and are mainly employed to represent the independent categorical data, e.g., channel indices for biological signals; (ii) level hypervectors are usually linearly correlated and represent the sub-intervals of a given range, e.g., the quantized magnitude of a given time series. The reader is referred to \cite{ge2020classification,heddes2022torchhd} for more details.

\begin{figure}[htbp] 
\centerline{\includegraphics[width=0.98\linewidth]{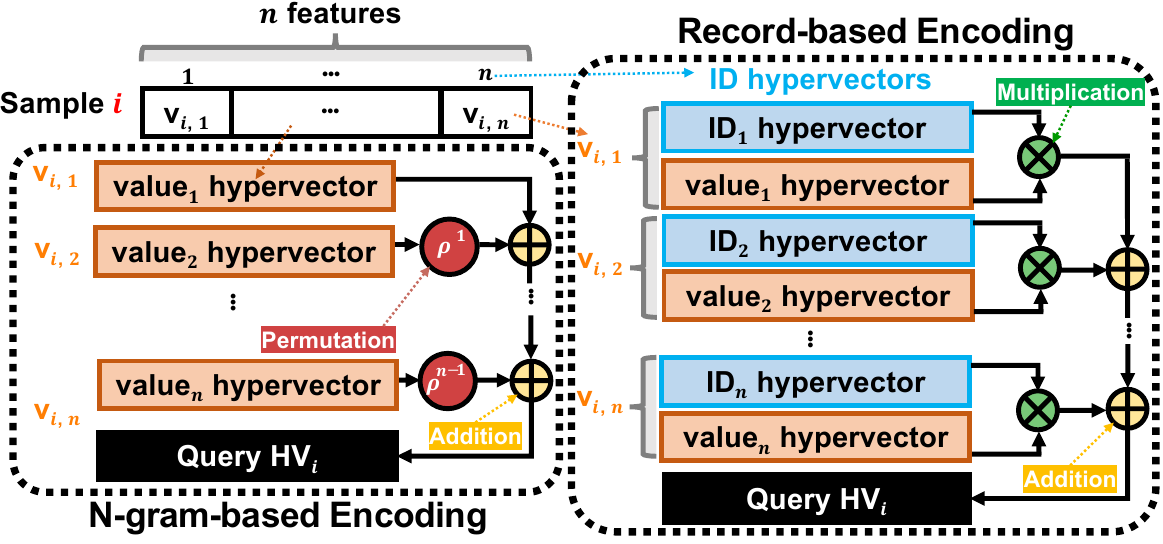}}
\caption{Two standard encoding algorithms in HDC.}\label{fig:stand_encoding}
\vspace{-12pt}
\end{figure}

\subsubsection{\textbf{Encoding onto Hyperdimensional Space}}
There are two standard encoding algorithms in the field of HDC to encode the original data onto the hyperdimensional space: record-based encoding and N-gram-based encoding.

\textbf{Record-based encoding \cite{ge2020classification}:} This encoding algorithm typically requires two types of hypervectors, which contain the value and position information, respectively. As shown in Fig. \ref{fig:stand_encoding}, the original given data, sample $i$, possesses $n$ different features. The $\mathbf{ID}$ hypervectors ($\{\mathbf{ID}_1, \cdots, \mathbf{ID}_n\}$) are used to represent the index of the feature, whereas the value $\mathbf{V}$ hypervectors ($\{\mathbf{V}_1, \cdots, \mathbf{V}_n\}$) reflect the feature value. Generally, the $\mathbf{ID}$ hypervectors are random hypervectors, and value hypervectors are level hypervectors whose quantization level is $q$. The encoded  $\mathbf{Query \, HV}_i$ hypervector is generated by Eq. (\ref{eq:rec}) in two steps: \textit{1).} Associate the feature value with its position by multiplication. \textit{2).} Add the calculated compound hypervectors in step 1. The bit-wise multiplication is denoted as $*$ in Eq. (\ref{eq:rec}). For binary HDC, the addition operation requires the majority rule to ensure the bit element of the hypervector is either 0 or 1.
\begin{equation}\label{eq:rec}
 \begin{aligned}
& \mathbf{Query \, HV}_i = \mathbf{\bar{V}}_1 * \mathbf{ID}_1 + \cdots + \mathbf{\bar{V}}_n * \mathbf{ID}_n,\\
& \mathbf{\bar{V}}_j  \in \{\mathbf{L}_1,\cdots, \mathbf{L}_q \}, \; \text{where} \; 1 \leq j \leq q.
\end{aligned}
\end{equation}

\textbf{N-gram-based encoding \cite{ge2020classification}:} The only difference between record-based encoding and N-gram-based encoding is how position and value are associated. Unlike record-based encoding, this algorithm encodes position information by permuting corresponding value hypervectors. For the $j^{th}$ index of the feature, the value hypervector $\mathbf{V}_j$ should be permuted ($j\!\!-\!\!1$) times. The whole encoding algorithm is described by Eq. (\ref{eq:ngram}).
\begin{equation}\label{eq:ngram}
 \begin{aligned}
& \mathbf{Query \, HV}_i= \mathbf{\bar{V}}_1 + \rho \mathbf{\bar{V}}_2 + \cdots + \rho ^{n-1} \mathbf{\bar{V}}_n,\\
& \mathbf{\bar{V}}_j  \in \{\mathbf{L}_1,\cdots, \mathbf{L}_q \}, \; \text{where} \; 1 \leq j \leq q.
\end{aligned}
\end{equation}

\subsubsection{\textbf{Similarity Measurement}}
In a nutshell, Hamming distance measures the similarity between binary hypervectors, whereas cosine similarity measures the similarity between non-binary hypervectors. A similarity measurement for two arbitrary hypervectors, $\delta(\mathbf{A}, \mathbf{B})$, can be calculated as shown in Eq. (\ref{eq:sim_check}), where $d$ is the dimensionality of $\mathbf{A}$ and $\mathbf{B}$.
\begin{equation}\label{eq:sim_check}
\resizebox{.9\hsize}{!}{$
 \delta(\mathbf{A},\mathbf{B})=
 \left\{
     \begin{array}{lr}
     \frac{1}{d}\sum_{i=1}^{d}1_{\mathbf{A}(i) \neq \mathbf{B}(i)}, \;\text{binary HDC},\\
     \frac{\mathbf{A} \cdot \mathbf{B}}{|\mathbf{A}||\mathbf{B}|}, \; \text{non-binary HDC},\\
    \end{array}
\right.$}
\end{equation}

In this paper, binary hypervectors refer to those hypervectors $\in \{0,1\}^d$, where non-binary hypervectors represent the bipolar seed hypervectors ($\in \{-1,1\}^d$) and compound hypervectors ($\in \{ \mathbb{Z} \}^d$) generated by bipolar hypervectors.


\begin{figure}[htbp] 
\centerline{\includegraphics[width=0.98\linewidth]{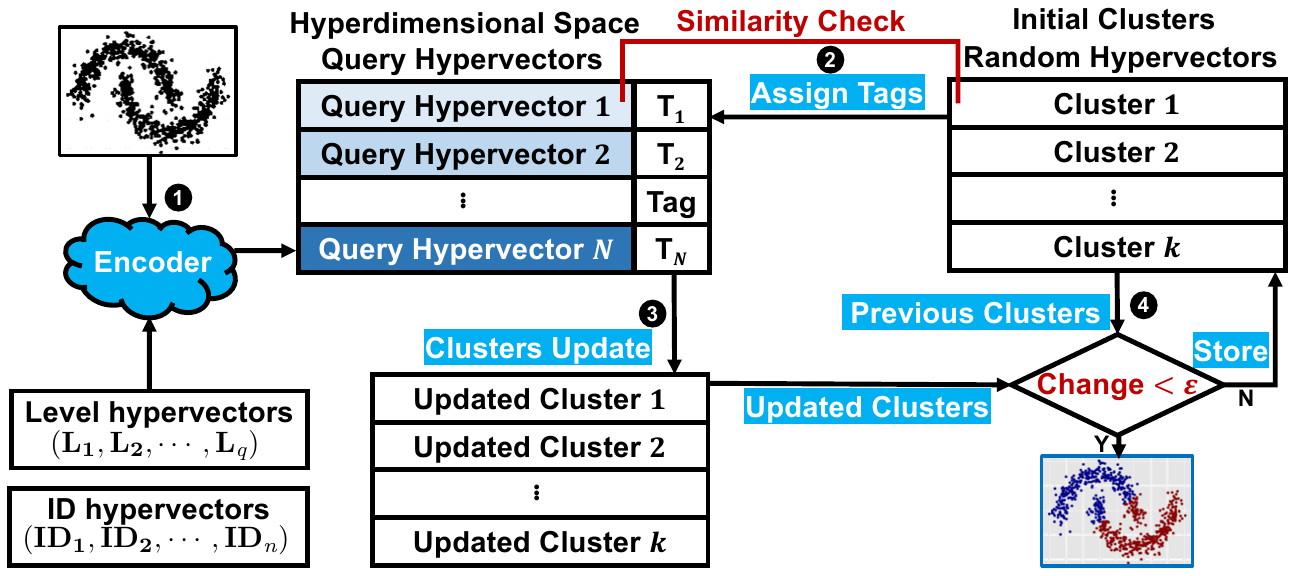}}
\caption{HDCluster overview \cite{imani2019hdcluster}. In \cite{imani2019hdcluster}, the encoder of the original HDCluster refers to record-based encoding, and the initial cluster centers are random hypervectors.}\label{fig:HDCluster_overview}
\vspace{-12pt}
\end{figure}

\begin{table*}[hbpt]
  \centering
  \setlength\tabcolsep{4pt}
  \caption{Datasets for Clustering using HDC.}
    \begin{adjustbox}{angle = 0, width=\textwidth}
    \begin{tabular}{c|c|c|c|c|c|c|c|c}
    \hline
    \rowcolor{black} \textcolor{white}{\textbf{Datasets}} & \textcolor{white}{\href{https://doi.org/10.24432/C53K8Q}{\textbf{MNIST}}} & \textcolor{white}{\href{https://doi.org/10.24432/C51G69}{\textbf{ISOLET}}} & \textcolor{white}{\href{https://doi.org/10.24432/C56C76}{\textbf{IRIS}}} & \textcolor{white}{\href{https://doi.org/10.24432/C5WW2P}{\textbf{Glass}}} & \textcolor{white}{\href{https://doi.org/10.24432/C5R88H}{\textbf{RNA-seq}}} & \textcolor{white}{\href{https://doi.org/10.24432/C5DW2B}{\textbf{Cancer}}} & \textcolor{white}{\href{https://doi.org/10.24432/C5388M}{\textbf{Ecoli}}} & \textcolor{white}{\href{https://doi.org/10.24432/C59C74}{\textbf{Parkinsons}}} \\
    \hline
    \# of Data Samples ($N$) & 10,000 & 7,797 & 150   & 214   & 801   & 569   & 336   & 195 \\
    \# of Features ($n$) & 784   & 617   & 4     & 9        & 20,531 & 30    & 7     & 22 \\
    \# of Clusters ($k$) & 10    & 26    & 3     & 6       & 5     & 2     & 8     & 2 \\
    \hline
    \end{tabular}\label{t:dataset}
    \end{adjustbox}
\end{table*}

\begin{figure*}[htbp] 
\centerline{\includegraphics[width=0.98\linewidth]{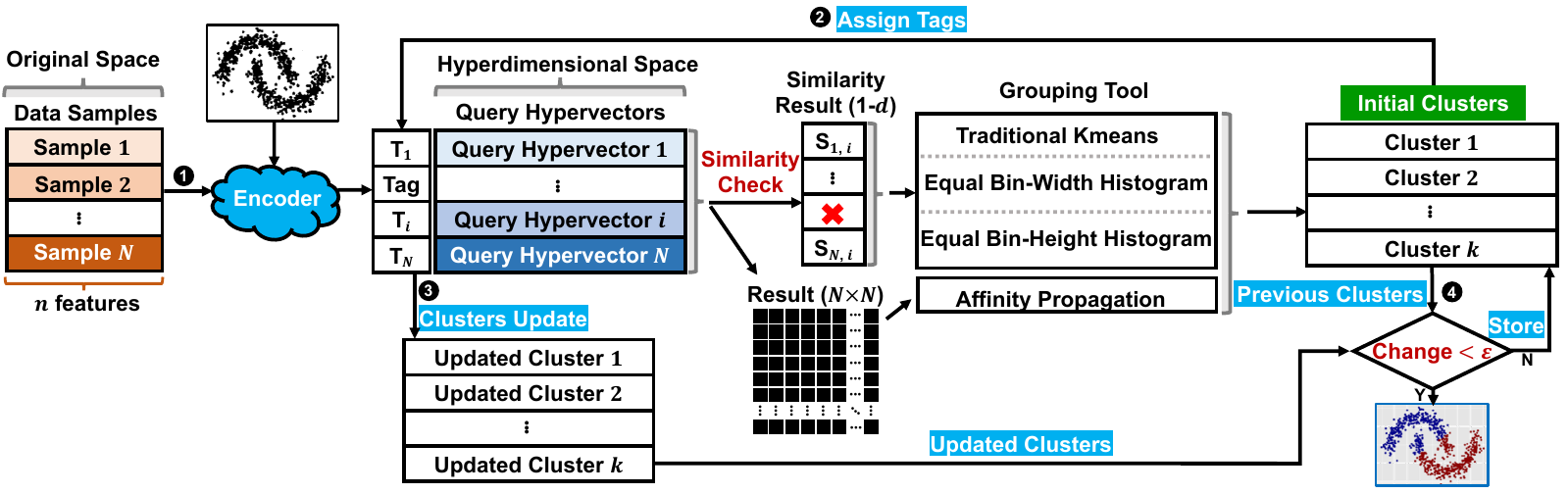}}
\caption{Proposed HDC-based clustering algorithms.}\label{fig:proposed_algorithm_overview}
\vspace{-12pt}
\end{figure*}

\subsection{HDCluster}\label{subsec:HDCluster}

HDCluster \cite{imani2019hdcluster} is a framework for HDC-based clustering, which is inspired by the traditional k-means clustering algorithm. Fig. \ref{fig:HDCluster_overview} illustrates an overview of HDCluster. For a $k$-cluster problem, the dataset contains $N$ data samples with $n$ features. At the very beginning, the initial $k$ cluster centers are assigned by $k$ random hypervectors. \ding{202} After the feature values are quantized into $q$ levels, each data sample is encoded as query hypervectors by the record-based encoding, where the feature indices are represented by $\mathbf{ID}$ hypervectors (random hypervectors), and the feature values are encoded by $\mathbf{V} \in \{\mathbf{L}_1, \cdots, \mathbf{L}_q\}$ (level hypervectors). \ding{203} In each iteration, each query hypervector is assigned to its cluster center which reflects the highest similarity, and is assigned a tag to represent the corresponding cluster. \ding{204} Cluster hypervectors are updated/regenerated by adding their associated query hypervectors with the same tag. \ding{205} The iterations will be terminated if (i) there is a minor change for the cluster hypervectors between two consecutive iterations or (ii) it exceeds the pre-defined number of iterations.

In the traditional k-means, the initial cluster centers are assigned from the data domain \cite{arthur2007k}. However, the assignment of cluster centers in HDCluster is from the data space ($\{0,1\}^d$). This assignment of initial cluster hypervectors in the HDCluster leads to non-robust clustering performance. More details are discussed in Sec. \ref{sec:results}.

\subsection{Dataset Description}
In this paper, we apply our HDC-based clustering algorithms to eight diverse datasets that are also tested in \cite{imani2019hdcluster}. The corresponding ground truth is provided. More details are described in Table \ref{t:dataset}.

\section{Methodology}\label{sec:method} 

In this section, we propose four novel HDC-based clustering algorithms. We assign all the $k$ initial cluster hypervectors from the data domain, in contrast to HDCluster which assigns these hypervectors based on the data space. To be more specific, the initial cluster hypervectors in this paper are no longer random hypervectors.

In HDC, original data points are encoded as hypervectors; thus, they are projected onto the hyperdimensional space. The relationships among those hypervectors are determined by their similarity measurement. Therefore, our HDC-based clustering algorithms are proposed based on the following assumptions: (i) the projection onto hyperdimensional space is helpful for data separability; (ii) the similarity measurement among hypervectors contains the information for clustering/grouping the data. 


To cluster the given $N$ data samples with our proposed HDC-based clustering algorithms below, the first three algorithms only require computing the similarity measurement $(N-1)$ times if we randomly pick a data sample as the starting point. However, the fourth algorithm needs to compute $N \choose 2$ similarity measurements.

\subsection{Similarity-Based K-means}
As shown in Fig. \ref{fig:proposed_algorithm_overview}, after the original data samples are encoded as query hypervectors, a starting point, i.e., sample query hypervector $i$, is randomly chosen among all $N$ data samples. Conduct the similarity measurement over all the other $(N-1)$ data samples with this starting point $i$, a 1-$d$ similarity result is obtained. A traditional k-means algorithm is then applied to this 1-$d$ similarity result to obtain the $k$ clusters.

\subsection{Equal Bin-Width Histogram}
\begin{figure}[H]
\centering
\vspace{-8pt}
\subfigure[Histogram with no zero-membered bins.]{
\includegraphics[width=0.45\linewidth]{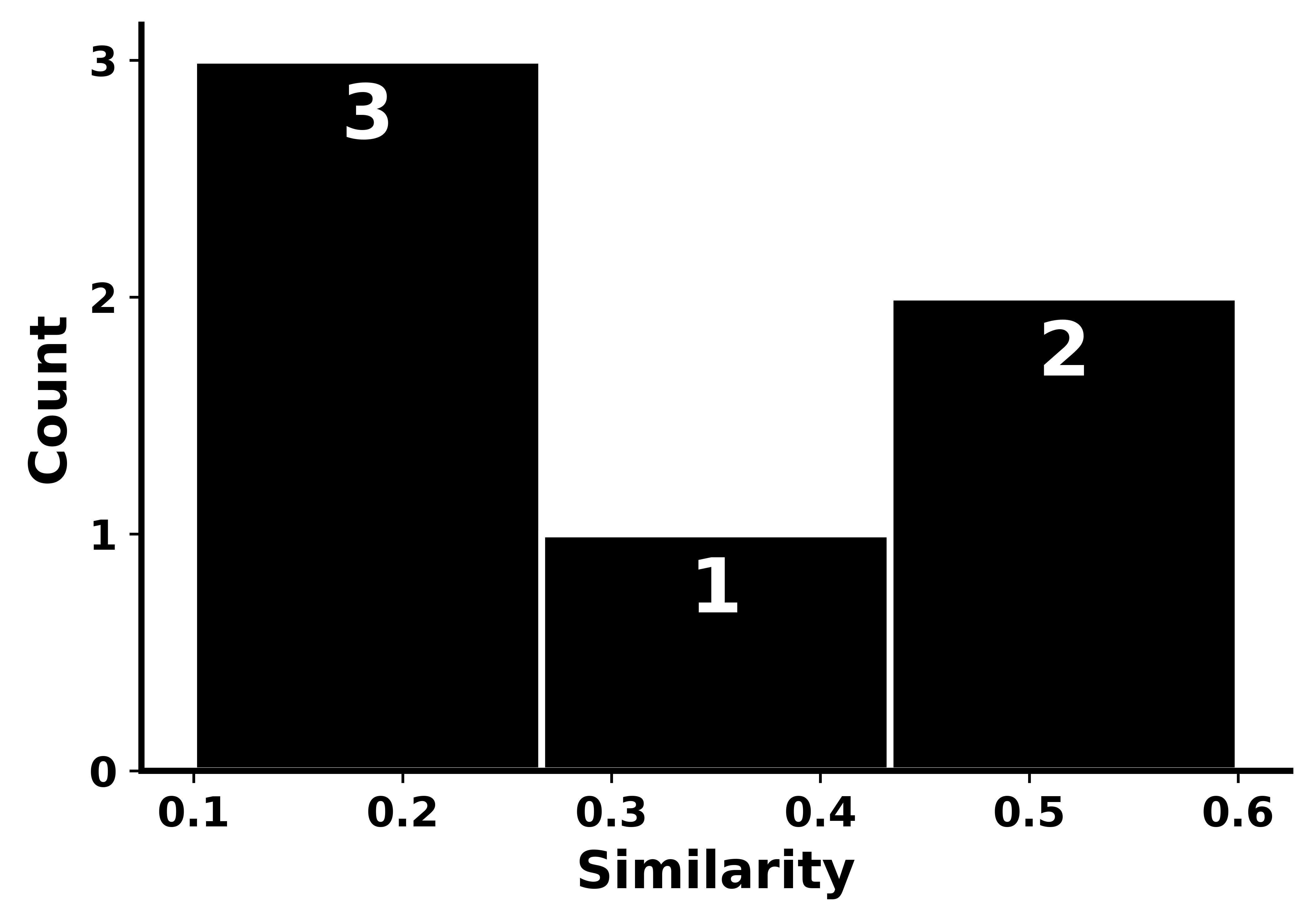}}\label{fig:hist_eq_bin_width_no_0}
\vspace{-8pt}
\subfigure[Histogram with zero-membered bins.]{
\includegraphics[width=0.45\linewidth]{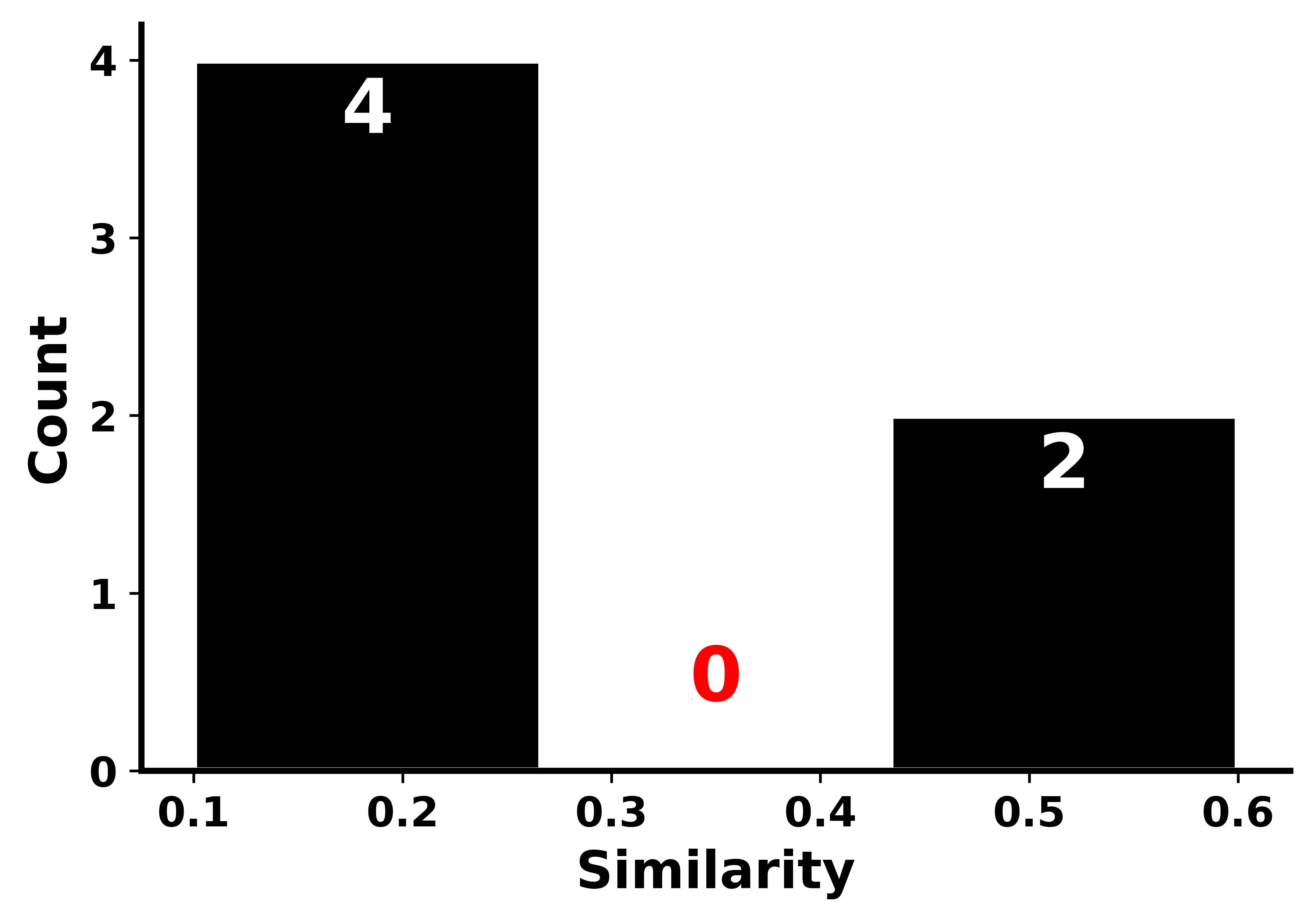}\label{fig:hist_eq_bin_width_wt_0}}
\caption{Examples for the equal bin-width histogram.}\label{fig:hist_eq_bin_width}
\vspace{-8pt}
\end{figure}

As a histogram is used to reflect the distribution of 1-$d$ data, it is natural to use a histogram to cluster or group data. The histogram is designed to have $k$ bins for any $k$ clustering problem. There are two typical styles of histograms: equal bin width and equal bin height. In this case, cluster hypervectors are evenly distributed in the hyperdimensional space between the data samples if a bin-width histogram is applied to the calculated 1-$d$ similarity result. Fig. \ref{fig:hist_eq_bin_width} shows two toy examples of six query hypervectors to be allocated into three clusters using the equal bin-width histogram. 

However, such a uniform distribution of cluster hypervectors cannot be guaranteed for any clustering problem. The histogram will contain some zero-membered bins in practice, e.g., Fig. \ref{fig:hist_eq_bin_width_wt_0}. We employ random hypervectors as the cluster hypervectors to represent these zero-membered bins. 

\subsection{Equal Bin-Height Histogram}
\begin{figure}[H]
\centering
\vspace{-8pt}
\subfigure[Divisible case.]{
\includegraphics[width=0.45\linewidth]{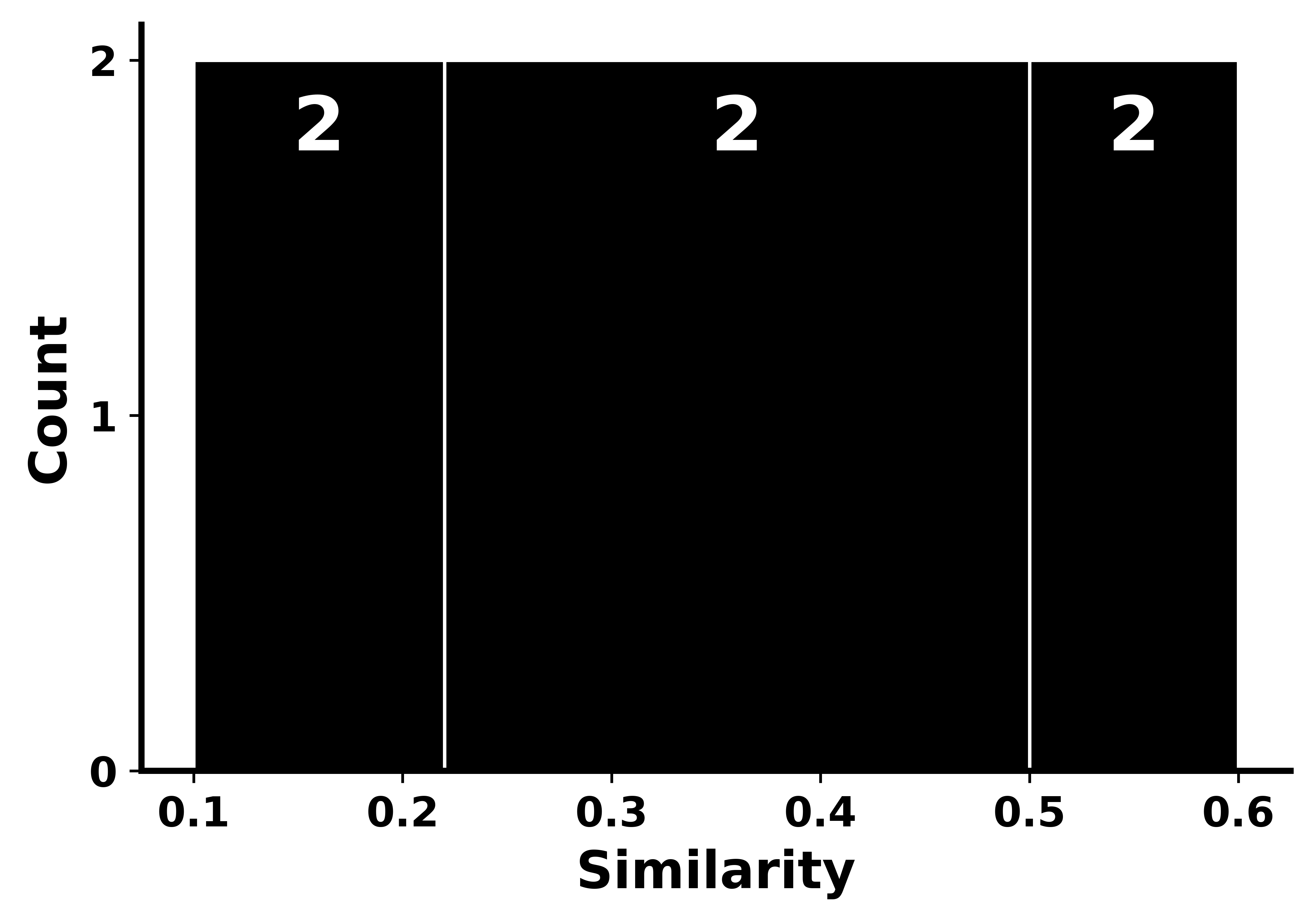}}\label{fig:hist_eq_bin_height_even}
\vspace{-8pt}
\subfigure[Indivisible case.]{
\includegraphics[width=0.45\linewidth]{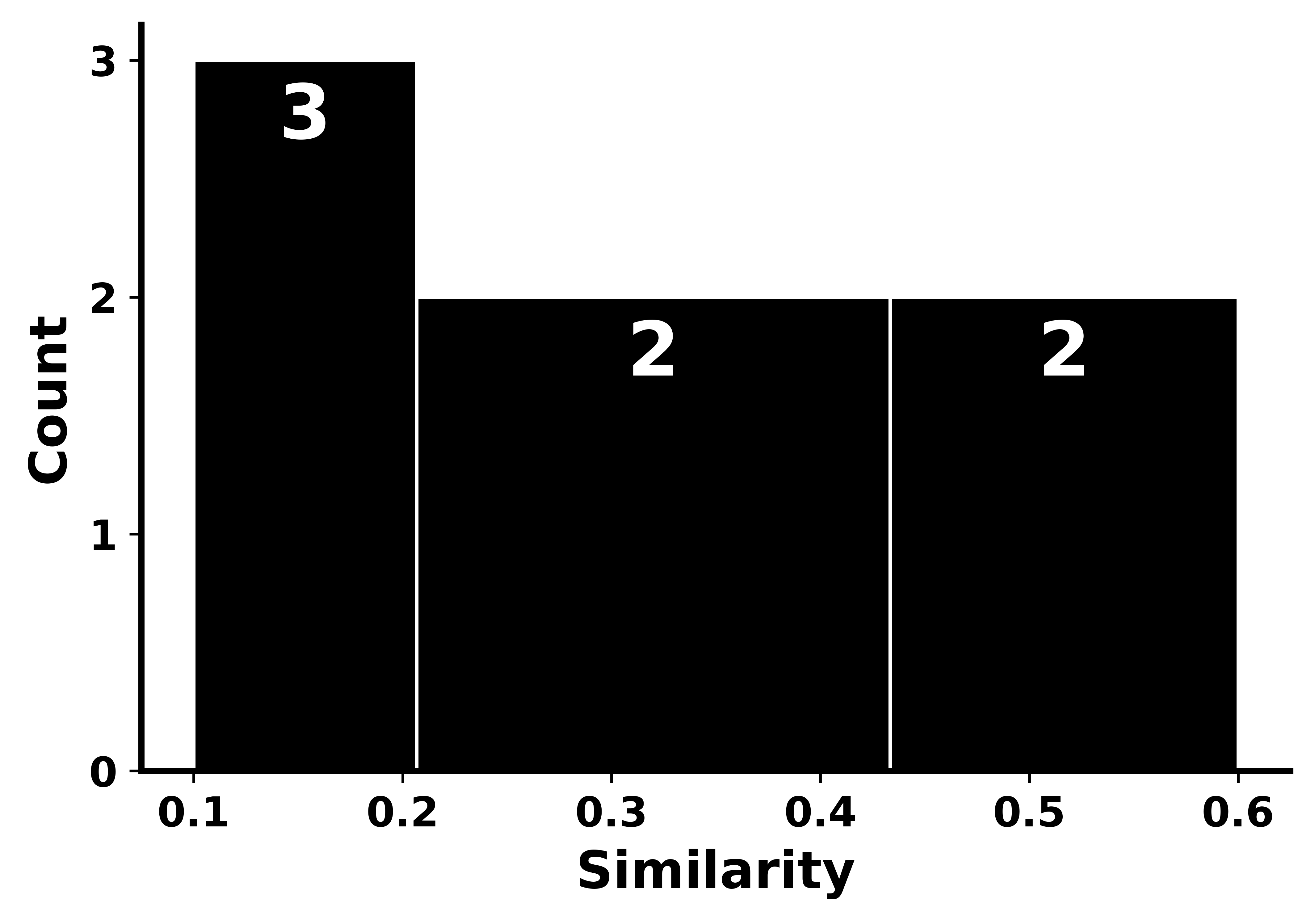}\label{fig:hist_eq_bin_height_odd}}
\caption{Examples for the equal bin-height histogram.}\label{fig:hist_eq_bin_height}
\vspace{-8pt}
\end{figure}

If an equal-height histogram is employed to the 1-$d$ similarity result, the corresponding assumption is that all these $k$ clusters have similar group sizes, i.e., the data are sampled from a uniform probability distribution. This equal bin-height histogram method is preferred when prior knowledge that the clusters have similar member sizes is given. Fig. \ref{fig:hist_eq_bin_height} gives two toy examples for the equal bin-height histogram. For the number of compared query hypervectors that can be divisible by the number of $k$ clusters ($\text{Mod}(\frac{N-1}{k})=0$, where Mod represents reminder), each cluster has the exact same group size, whereas the group sizes of $k$ clusters are roughly similar for the indivisible case ($\text{Mod}(\frac{N-1}{k})\neq 0$).

\subsection{Similarity-Based Affinity Propagation} 
Different from the previous three algorithms, this algorithm requires an $N \times N$ similarity result. Our similarity-based affinity propagation is a variant of traditional affinity propagation. Different from traditional affinity propagation whose similarity matrix is typically measured by the pairwise Euclidean distance of the raw data, our proposed similarity-based affinity propagation starts with the similarity result measured by Eq. (\ref{eq:sim_check}) for the encoded data (query hypervectors). The similarity result for hypervectors is fed into the traditional affinity propagation algorithm \cite{frey2007clustering} to obtain the clustering results. 


\section{Experimental Results}\label{sec:results}  

\subsection{Experimental Setup}

We implement both HDCluster and our proposed HDC-based clustering algorithms using Python implementation. For hypervectors generation, we employ the library ``torchhd'' \cite{heddes2022torchhd}. We evaluate the clustering algorithms by three metrics: accuracy (ground truth is known), number of iterations for the convergence of cluster hypervectors, and execution time. As mentioned above, the generation of random hypervectors has an impact on the clustering performance of HDCluster. To capture the non-robust performance, we test all the HDC-based clustering algorithms over $500$ runs with different pseudo seeds. Thus in Python code, the ``torch.manual\_seed'' ranges from 0 to 499. This enables us to compute the variance of the performance results. Note that, in HDCluster, the algorithm was run only once and no variance was reported.

\begin{figure*}[ht]
\centering
\vspace{-8pt}
\includegraphics[width=\linewidth]{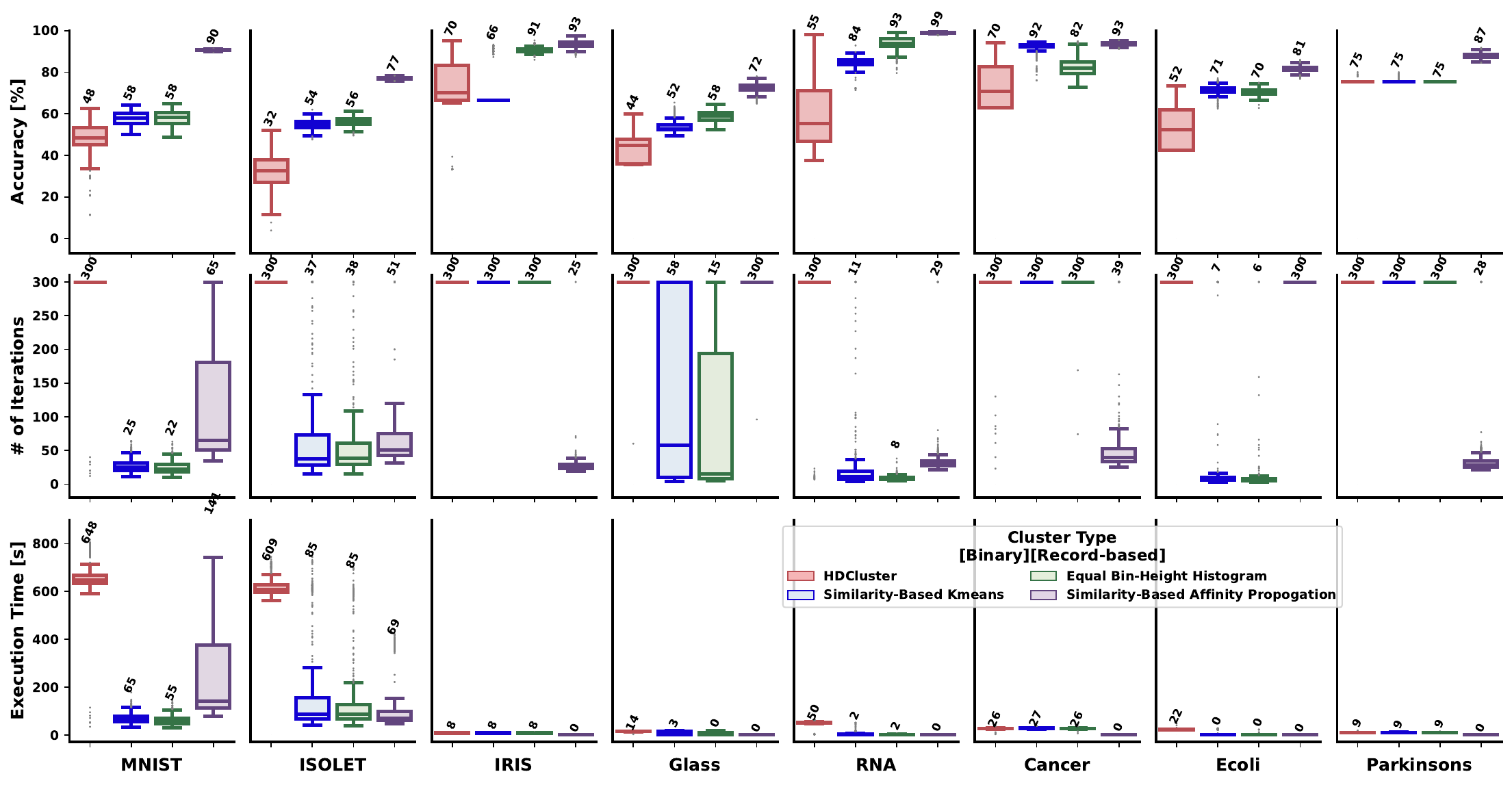}\label{fig:per_vs_imani_ham_re}
\caption{Comparison of proposed algorithms with HDCluster in accuracy (top row), number of iterations (middle row), and execution time (bottom row) using binary HDC with record-based encoding. For boxplots, the median values over 500 runs are annotated on the top.}\label{fig:performace_vs_imani}
\vspace{-8pt}
\end{figure*}

The projection onto hyperdimensional space requires the quantization of the original data. As with \cite{imani2019hdcluster}, quantization level $q$ is set to $16$ to ensure a fair comparison. Additionally, the dimensionality of hypervectors is set as $d=10,000$ for all experiments.

To measure the \textit{minor} change of cluster hypervectors between two consecutive iterations, the minimum cosine similarity (over $k$) between current ($\{\mathbf{C_1}', \cdots, \mathbf{C_k}'\}$) and previous ($\{\mathbf{C_1}, \cdots, \mathbf{C_k}\}$) cluster hypervectors should be greater than $0.99$ for non-binary HDC, whereas for the maximum Hamming distance between current and previous cluster hypervectors should be less than $0.01$ for binary HDC. The maximum number of iterations for termination of the iterative update is predefined as $300$ in this paper.

Note in \cite{imani2019hdcluster}, both the seed hypervectors and the compound hypervectors are binary hypervectors ($\in \{0,1\}^d$). In this paper, similar to \cite{imani2019hdcluster}, we test our proposed HDC-based algorithms using binary hypervectors. We also apply our algorithms using non-binary HDC, which means that seed hypervectors are bipolar and compound hypervectors are integers.

\subsection{Comparison with HDCluster}

Using HDCluster as a baseline framework, we test our proposed HDC-based clustering algorithms over eight datasets. Two encoding algorithms, record-based and N-gram-based encoding, are employed. Both binary HDC and non-binary HDC are examined. Additionally, we also compare algorithms using one-pass clustering with HDCluster.

\begin{table*}[hbpt]
  \centering
  \caption{Clustering performance over 500 runs using Binary HDC with record-based encoding.}
    \begin{adjustbox}{angle = 0, width=\textwidth}
    \begin{tabular}{c|c|c|c|c|c|c|c|c}
   \noalign{\hrule height 1pt}
\rowcolor{black}     \textcolor{white}{\textbf{Method}} & \textcolor{white}{\href{https://doi.org/10.24432/C53K8Q}{\textbf{MNIST}}} & \textcolor{white}{\href{https://doi.org/10.24432/C51G69}{\textbf{ISOLET}}} & \textcolor{white}{\href{https://doi.org/10.24432/C56C76}{\textbf{IRIS}}} & \textcolor{white}{\href{https://doi.org/10.24432/C5WW2P}{\textbf{Glass}}} & \textcolor{white}{\href{https://doi.org/10.24432/C5R88H}{\textbf{RNA-seq}}} & \textcolor{white}{\href{https://doi.org/10.24432/C5DW2B}{\textbf{Cancer}}} & \textcolor{white}{\href{https://doi.org/10.24432/C5388M}{\textbf{Ecoli}}} & \textcolor{white}{\href{https://doi.org/10.24432/C59C74}{\textbf{Parkinsons}}} \\
  \noalign{\hrule height 1pt} 
   \multicolumn{9}{c}{\textcolor{black}{\textbf{Accuracy [$\%$]}}} \\
      \noalign{\hrule height 1pt}
\textbf{HDCluster}    & 47.10($\pm$8.79) & 32.26($\pm$8.11) & 72.39($\pm$13.85) & 43.17($\pm$6.07) & 59.01($\pm$16.05) & 73.50($\pm$10.44) & 53.49($\pm$10.80) & 75.45($\pm$0.47) \\
     \textbf{SB Kmeans}    & 57.87($\pm$3.35) & 54.80($\pm$2.13) & 72.52($\pm$10.39) & 53.97($\pm$2.73) & 84.67($\pm$2.41) & 92.17($\pm$2.31) & 71.30($\pm$2.04) & 75.60($\pm$0.78) \\
   \textbf{Bin Height}  & 57.96($\pm$3.36) & 56.33($\pm$1.94) & 90.96($\pm$1.35) & 58.81($\pm$2.40) & 93.70($\pm$3.37) & 82.52($\pm$4.73) & 70.50($\pm$1.60) & 75.38($\pm$0.00) \\
    \textbf{SB Affinity Propagation}   & 90.67($\pm$0.26) & 77.07($\pm$0.46) & 93.37($\pm$1.76) & 72.36($\pm$2.37) & 99.03($\pm$0.34) & 93.55($\pm$0.69) & 81.65($\pm$1.45) & 87.92($\pm$1.24) \\
   \noalign{\hrule height 1pt} 
   \multicolumn{9}{c}{\textbf{Number of Iterations}} \\
      \noalign{\hrule height 1pt}
  \textbf{HDCluster}     & 295.61($\pm$34.51) & 300.00($\pm$0.00) & 300.00($\pm$0.00) & 299.52($\pm$10.73) & 289.05($\pm$55.18) & 296.40($\pm$28.56) & 300.00($\pm$0.00) & 300.00($\pm$0.00)   \\
     \textbf{SB Kmeans}     & 26.83($\pm$9.35) & 84.77($\pm$98.57) & 300.00($\pm$0.00) & 141.94($\pm$137.40) & 22.73($\pm$45.16) & 300.00($\pm$0.00) & 62.93($\pm$114.26) & 300.00($\pm$0.00)   \\
    \textbf{Bin Height}     & 24.25($\pm$9.10) & 71.54($\pm$82.56) & 300.00($\pm$0.00) & 91.61($\pm$120.86) & 8.86($\pm$3.32) & 299.29($\pm$11.67) & 10.01($\pm$28.03) & 300.00($\pm$0.00)   \\
     \textbf{SB Affinity Propagation}    & 120.65($\pm$104.16) & 94.82($\pm$94.32) & 27.79($\pm$18.18) & 299.59($\pm$9.12) & 37.20($\pm$40.10) & 64.54($\pm$73.24) & 300.00($\pm$0.00) & 37.72($\pm$43.68)   \\
   \noalign{\hrule height 1pt} 
   \multicolumn{9}{c}{\textcolor{white}{\textbf{Execution Time [s]}}} \\
      \noalign{\hrule height 1pt}
    \textbf{HDCluster}     & 655.90($\pm$90.51) & 620.11($\pm$40.00) & 8.41($\pm$0.79) & 15.11($\pm$1.31) & 50.08($\pm$10.36) & 27.58($\pm$3.79) & 23.35($\pm$2.29) & 9.53($\pm$1.08) \\
     \textbf{SB Kmeans}     & 69.26($\pm$21.10) & 184.01($\pm$204.09) & 8.61($\pm$0.80) & 7.44($\pm$7.01) & 4.67($\pm$7.56) & 28.36($\pm$2.79) & 5.13($\pm$8.75) & 9.78($\pm$1.01) \\
    \textbf{Bin Height}     & 61.13($\pm$20.40) & 151.84($\pm$165.57) & 8.36($\pm$0.78) & 4.72($\pm$5.97) & 2.22($\pm$0.60) & 27.69($\pm$2.97) & 1.07($\pm$2.07) & 9.50($\pm$1.03) \\
   \textbf{SB Affinity Propagation}   & 255.15($\pm$211.93) & 124.40($\pm$117.13) & 0.02($\pm$0.01) & 0.22($\pm$0.01) & 0.36($\pm$0.32) & 0.29($\pm$0.28) & 0.45($\pm$0.02) & 0.03($\pm$0.03) \\
    \noalign{\hrule height 1pt}
    \end{tabular}\label{t:binary_re_perf_summary}
    \end{adjustbox}
\end{table*}

\subsubsection{\textbf{Iterative Update of the Center Hypervectors}}
Fig. \ref{fig:performace_vs_imani} shows the boxplots for the experimental results of our proposed HDC-based clustering algorithms and the baseline HDCluster over 500 runs using binary HDC with record-based encoding. Median values are annotated on the top of the boxplots. As shown in Fig. \ref{fig:performace_vs_imani}, our proposed algorithms are more robust in clustering accuracy, require fewer iterative updates in cluster hypervectors, and consume less execution time over all eight datasets as compared to HDCluster. To be more specific, \textbf{(i) Accuracy}: Among our proposed HDC-based clustering algorithms, similarity-based affinity propagation always achieves the highest clustering accuracy over all eight datasets.  \textbf{(ii) The number of iterations}: HDCluster reaches the maximum pre-specified iteration value (=300) more often than our proposed four algorithms.  \textbf{(iii) Execution time}: HDCluster nearly always requires a longer execution time---especially for MNIST and ISOLET datasets---in comparison to our proposed algorithms. This validates that, by randomly assigning the initial clusters from the hyperdimensional space, the existing HDCluster can not perform a fast clustering for a dataset with large data samples (e.g., MNIST: 10,000, ISOLET: 7,797).


\begin{figure*}[!htbp]
\centering
{\includegraphics[width=\linewidth]{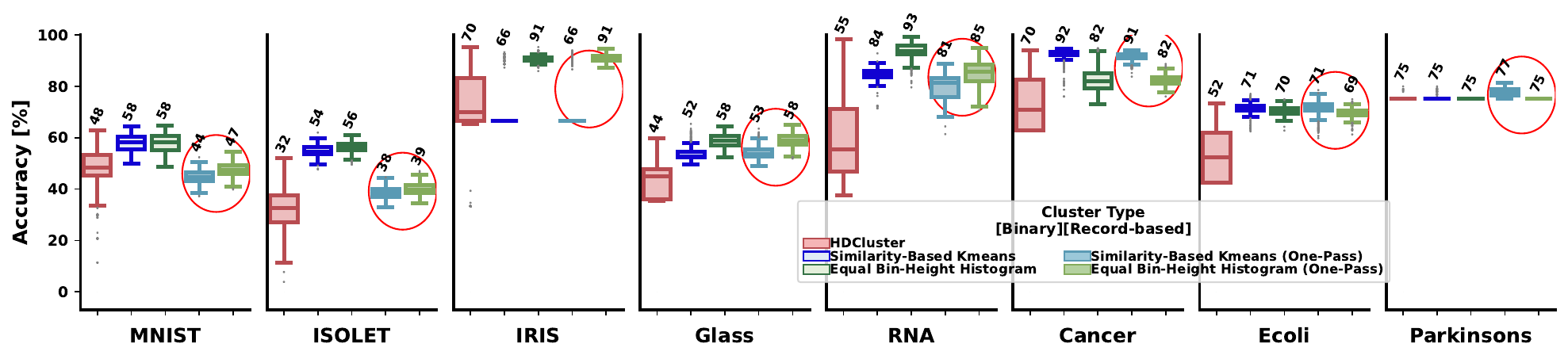}}\label{fig:1_shot_ham_re}
\caption{One-pass clustering performance of our proposed algorithms as compared to HDCluster and our iterative algorithms using binary HDC with record-based encoding. For boxplots, the median values over 500 runs are annotated on the top. The red circles highlight our one-pass clustering results. }\label{fig:performance_1shot}
\end{figure*}
\begin{figure*}[ht]
\centering
{\includegraphics[width=\linewidth]{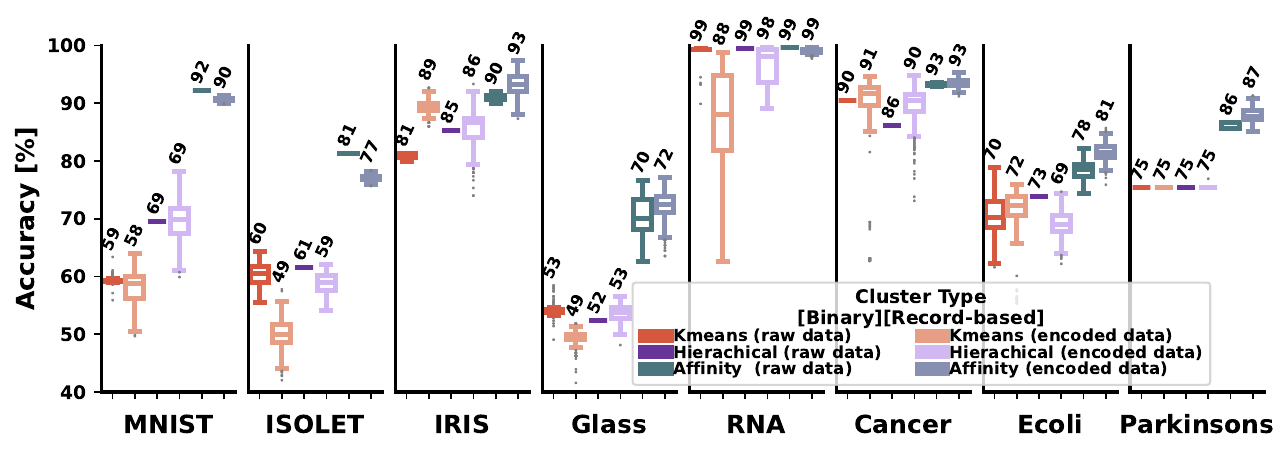}}\label{fig:query_ham_re}
\caption{Comparison of clustering using original data and encoded data for binary HDC with record-based encoding. For boxplots, the median values over 500 runs are annotated on the top. }\label{fig:performance_vs_query}
\vspace{-8pt}
\end{figure*}
\subsubsection{\textbf{One-Pass Clustering}}
One-pass clustering refers to a single pass clustering without any additional iterations. To be more specific, here one-pass clustering only involves \ding{202}-\ding{203} of Fig. \ref{fig:proposed_algorithm_overview}. Without any doubt, one-pass clustering requires less execution time and fewer iterations for convergence as compared to the iterative update of the cluster hypervectors. For comparison purposes, only accuracy is considered. Note that similarity-based affinity propagation is not considered for a one-pass clustering scenario.




The experimental results using binary HDC with record-based encoding for one-pass clustering are shown in Fig. \ref{fig:performance_1shot}. Regarding the clustering accuracy, the one-pass clustering algorithms are lower than or comparable to the iterative update version of the proposed algorithms for four datasets (MNIST, ISOLET, RNA, and Cancer), while they can achieve comparable performance for the other four datasets. Additionally, compared to HDCluster, the clustering accuracy for our one-pass algorithms is also more robust. As shown by the median values for the boxplots, our one-pass clustering algorithms are even more likely to be more accurate than HDCluster.

\subsection{Original Data Space vs Hyperdimensional Space}\label{subsub:original_vs_hd}
We now address the question of whether projection onto hyperdimensional space is helpful for data separation. We employ traditional k-means, hierarchical clustering, and affinity propagation toward both the original data and the encoded data (query hypervectors). Both binary and non-binary HDC, associated with record-based encoding and N-gram-based encoding, are employed. Therefore, all four possible cases are examined: binary/non-binary HDC with record-based/N-gram-based encoding. The corresponding experimental results using binary HDC with record-based encoding are shown in Fig. \ref{fig:performance_vs_query}.


\begin{table*}[!hbpt]
  \centering
  \setlength\tabcolsep{1.3pt}
  \caption{Performance using the projected data for three traditional clustering algorithms$^*$.}
  \vspace{-4pt}
    \begin{tabular}{c|c|c|c|c|c|c|c|c}
    \hline
     \rowcolor{black} \textcolor{white}{\textbf{Datasets}} & \textcolor{white}{\href{https://doi.org/10.24432/C53K8Q}{\textbf{MNIST}}} & \textcolor{white}{\href{https://doi.org/10.24432/C51G69}{\textbf{ISOLET}}} & \textcolor{white}{\href{https://doi.org/10.24432/C56C76}{\textbf{IRIS}}} & \textcolor{white}{\href{https://doi.org/10.24432/C5WW2P}{\textbf{Glass}}} & \textcolor{white}{\href{https://doi.org/10.24432/C5R88H}{\textbf{RNA-seq}}} & \textcolor{white}{\href{https://doi.org/10.24432/C5DW2B}{\textbf{Cancer}}} & \textcolor{white}{\href{https://doi.org/10.24432/C5388M}{\textbf{Ecoli}}} & \textcolor{white}{\href{https://doi.org/10.24432/C59C74}{\textbf{Parkinsons}}} \\
    \hline
    \# of Data Samples ($N$) & 10,000 & 7,797 & 150   & 214   & 801   & 569   & 336   & 195 \\
    \# of Features ($n$) & 784   & 617   & 4     & 9        & 20,531 & 30    & 7     & 22 \\
    \# of Clusters ($k$) & 10    & 26    & 3     & 6       & 5     & 2     & 8     & 2 \\
    \hline
    Projection for k-means &  \ding{52}   & \textcolor{orange}{\ding{56}}    &  \textcolor{greenish}{\ding{52}}  &   \textcolor{orange}{\ding{56}}       &   \textcolor{orange}{\ding{56}}   &  \textcolor{greenish}{\ding{52}}  &   \ding{52}  &  \ding{52}\\
    Projection for Hierarchical Clustering  & \ding{52}   & \textcolor{orange}{\ding{56}}   &  \textcolor{greenish}{\ding{52}}  &   \ding{52}   &     \ding{52}  & \textcolor{greenish}{\ding{52}} &   \textcolor{orange}{\ding{56}}   &  \ding{52} \\
    Projection for Affinity Propagation & \textcolor{orange}{\ding{56}}    & \textcolor{orange}{\ding{56}}    &  \textcolor{greenish}{\ding{52}}  &   \ding{52}     &   \ding{52}  & \textcolor{greenish}{\ding{52}}  &   \ding{52}  &  \textcolor{greenish}{\ding{52}} \\
    \hline
    \multicolumn{9}{l}{\specialcell[b]{$^*$In last three rows, accuracy performance compared to original data: comparable (\ding{52}),  significantly better (\textcolor{greenish}{\ding{52}}), and lower (\textcolor{orange}{\ding{56}}).}}
    \end{tabular}\label{t:projection_or_not}
     \vspace{-4pt}
\end{table*} 

From Fig. \ref{fig:performance_vs_query}, the traditional clustering approach is better for at most three out of eight datasets, compared to HDC. To be more specific, \textit{1).} \textbf{Traditional k-means:} no projection is preferred for three datasets: ISOLET, RNA, and Glass. \textit{2).} \textbf{Hierarchical Clustering:} Both ISOLET and Ecoli datasets do not benefit from projection onto hyperdimensional space, as this may cause $2\%\sim4\%$ accuracy degradation. \textit{3).} \textbf{Affinity Propagation:} MNIST and ISOLET datasets experience an accuracy drop (of approximately $2\%$ and $5\%$, respectively) for hyperdimensional projection.

\subsection{Further Discussion}

\subsubsection{\textbf{The High/Robust Accuracy and Fast Convergence of Our Proposed HDC-based Clustering Algorithms}} As mentioned in Sec. \ref{subsec:HDCluster}, the initial cluster hypervectors of HDCluster are random seed hypervectors, that are either ($\{0,1\}^d$) or ($\{-1,1\}^d$). It indicates there exist in total $2^d \choose k$ ways of initializing the $k$ cluster hypervectors, which leads to the non-robust clustering accuracy performance. Low accuracy is easily obtained if the positions of the initially assigned cluster hypervectors in the hyperdimensional space are all far away from the encoded query hypervectors, so that query hypervectors have the same similarity results as the cluster hypervectors and cannot be correctly separated/clustered. High accuracy could be achieved when the initially assigned $k$ clusters have different similarity results with the query hypervectors. Our proposed HDC-based clustering algorithms assign the initial cluster hypervectors from the data domain. In other words, our algorithms utilize the information leveraged by the query hypervectors. As a result, the initial $k$ cluster hypervectors are determined by query hypervectors in our algorithms and the source of the accuracy variance over $500$ runs only comes from the inevitable randomness of seed hypervectors. Additionally, our data-domain-based assignment speeds up the convergence for the iterative update of cluster hypervectors. 

Based on Figs. \ref{fig:performace_vs_imani} and \ref{fig:performace_vs_imani(conti.)}, we find the similarity-based affinity propagation always achieves the highest clustering accuracy for binay/non-binary HDC with record-/N-gram-based encoding over all eight datasets. Particularly, 
there exists a significant improvement by this similarity-based affinity propagation for MNIST ($\approx\!\!38\%$), ISOLET ($\approx\!\!24\%$), Glass ($\approx\!\!17\%$), Ecoli ($\approx\!9\%$), and Parkinson's ($\approx\!9\%$), as compared to the other HDC-based clustering algorithms. This finding can also be observed from Table \ref{t:app:mean_std_ours_vs_hdcluster}.

\subsubsection{\textbf{Projection onto Hyperdimensional Space}} Table \ref{t:projection_or_not} summarizes the comparison of original data with encoded hypervectors across all four scenarios (two types of HDC and two encoding algorithms). To summarize, the projection of five datasets onto hyperdimensional space can lead to comparable or better accuracy performance compared to traditional k-means, hierarchical clustering, and affinity propagation. For all of these three traditional clustering algorithms, ISOLET performance is better for the original space, while IRIS performance is better for the hyperdimensional space. This indicates that projection onto hyperdimensional space might not be helpful for data separability when we have a large number of clusters, e.g., $k = 26$. Additionally, the projection onto hyperdimensional space can lead to higher performance when the number of clusters $k$ and the number of features $n$ are both small.

\subsubsection{\textbf{Similarity-based Hierarchical Clustering is Excluded in Our Proposed HDC-based Clustering Algorithms}} Similar to the similarity-based affinity propagation, we also feed the $N \times N$ similarity result into the hierarchical clustering. The corresponding performance is significantly lower than that of both the raw data and encoded data for affinity propagation (Sec. \ref{subsub:original_vs_hd}). As a result, similarity-based hierarchical clustering is not considered in our algorithms. 



\subsubsection{\textbf{Applications of HDC-based Clustering}}
Several recent works \cite{imani2018hdna,zou2022biohd} indicate a trend of applying HDC to biological datasets that can obtain surprisingly great performance. Based on our experimental results over eight datasets (Figs. \ref{fig:performace_vs_imani} and \ref{fig:performance_vs_query}), HDC-based clustering can achieve high accuracy ($>\!90\%$) in RNA and Cancer datasets. Therefore, the combination of HDC with biological datasets could provide interesting results and new insights.

\balance
\section{Conclusion}\label{sec:con}
We demonstrate that the existing HDCluster suffers from non-robust cluster accuracy and a large number of convergence iterations as a consequence of the data-space-based assignment of the initial cluster hypervectors.
We propose four HDC-based clustering algorithms based on the categorized information that take advantage of the encoded data---intra-cluster hypervectors have a higher similarity than inter-cluster hypervectors. We measure our algorithms by employing two standard encoding algorithms (record-based and N-gram-based encoding) for both binary and non-binary HDC. As compared to the existing HDCluster, our proposed HDC-based algorithms achieve better and more robust accuracy, fewer iterative updates of cluster hypervectors, and less execution time when tested over eight datasets. Similarity-based affinity propagation outperformed the other three HDC-based clustering algorithms on eight datasets by $ 2\%\sim38\%$ in clustering accuracy. Even for one-pass clustering, our proposed algorithms can provide more robust clustering accuracy than HDCluster. In terms of whether to use the original data or encoded data, we find that five out of eight datasets that are projected onto hyperdimensional space can achieve better or comparable clustering accuracy as compared to the original space. In particular, ISOLET does not require projection, whereas IRIS benefits from hyperdimensional projection. This observation implies that projecting onto hyperdimensional space is attractive when both the number of clusters $k$ and the number of features $n$ are small. Maintaining the original data space is recommended when the number of clusters, $k$, is large. Future work will be directed towards three avenues. First, all the discussed clustering problems in this paper are provided with ground truth. Additionally, the target number of clusters $k$ is already known. The capability of HDC to infer the optimal number of clusters $k$ from a given dataset without ground truth should be investigated. Second, the algorithms in this paper are software-oriented. Due to HDC's energy efficiency, future work should be directed toward hardware implementations. Third, we find that HDC-based clustering algorithms perform well in RNA and Cancer datasets. Therefore, future efforts should address clustering different types of biological datasets in the HDC domain to obtain better performance and gain new insights.

\appendices
\section{Statistics of the clustering accuracy for HDCluster and our proposed HDC-based algorithms}

Table \ref{t:app:mean_std_ours_vs_hdcluster} summarizes the mean and standard deviation results of the clustering accuracy over 500 runs for both the baseline HDCluster and our algorithms. Our proposed algorithms are tested for two types of HDC (binary HDC and non-binary HDC) and two encoding algorithms (record-based and N-gram-based encodings). The standard deviation reflects the variance of the accuracy. For each dataset, the highest performance is in bold.

\begin{table*}[t]
  \centering
  \setlength\tabcolsep{0.2pt}
  \caption{Performance comparison of our proposed algorithms and the baseline HDCluster$^*$.}
    \begin{adjustbox}{angle = 0, width=\textwidth}
    \begin{tabular}{@{}l!{\vrule width 0.9pt}cccccccc@{}}
    \noalign{\hrule height 1pt}
    \multicolumn{1}{c!{\vrule width 0.9pt}}{\textbf{Method$^{**}$}} & \multicolumn{1}{c}{\textbf{MNIST}} & \multicolumn{1}{c}{\textbf{ISOLET}} & \multicolumn{1}{c}{\textbf{RNA}}   & \multicolumn{1}{c}{\textbf{Cancer}} & \multicolumn{1}{c}{\textbf{IRIS}}  & \multicolumn{1}{c}{\textbf{Glass}} & \multicolumn{1}{c}{\textbf{Ecoli}} & \multicolumn{1}{c}{\textbf{Parkinson's}} \\
    \noalign{\hrule height 1pt}
    \multicolumn{9}{c}{\textbf{Binary HDC using Record-based Encoding}} \\
    \noalign{\hrule height 1pt}
    \textbf{HDCluster} & 47.10($\pm$8.79) & 32.26($\pm$8.11) & 59.01($\pm$16.05) & 73.50($\pm$10.44) & 72.39($\pm$13.85) & 43.17($\pm$6.07) & 53.49($\pm$10.80) & 75.45($\pm$0.47) \\
    \textbf{SB Kmeans} & 57.87($\pm$3.35) & 54.80($\pm$2.13) & 84.67($\pm$2.41) & 92.17($\pm$2.31) & 72.52($\pm$10.39) & 53.97($\pm$2.73) & 71.30($\pm$2.04) & 75.60($\pm$0.78) \\
    \textbf{Bin Height} & 57.96($\pm$3.36) & 56.33($\pm$1.94) & 93.70($\pm$3.37) & 82.52($\pm$4.73) & 90.96($\pm$1.35) & 58.81($\pm$2.40) & 70.50($\pm$1.60) & 75.38($\pm$0.00) \\
    \textbf{Bin Width}  & / & / & / & 86.52($\pm$2.54) & 66.67($\pm$0.00) & 52.34($\pm$0.94) & / & 75.38($\pm$0.00) \\
    \hline
    \specialcell[c]{\textbf{SB Kmeans (1-Pass)}} & 44.70($\pm$2.34) & 38.60($\pm$2.17) & 79.75($\pm$4.86) & 91.57($\pm$1.52) & 72.21($\pm$10.18) & 54.26($\pm$2.42) & 71.43($\pm$2.74) & 77.59($\pm$1.55) \\
    \specialcell[c]{\textbf{Bin Height (1-Pass)}} & 47.58($\pm$2.59) & 39.94($\pm$2.20) & 85.24($\pm$4.68) & 82.32($\pm$1.87) & 91.06($\pm$1.28) & 58.93($\pm$2.39) & 69.54($\pm$1.72) & 75.38($\pm$0.00) \\
    \specialcell[c]{\textbf{Bin Width (1-Pass)}} & / & / & / & 73.19($\pm$1.05) & 66.67($\pm$0.00) & 51.73($\pm$0.79) & / & 75.38($\pm$0.00) \\
    \hline
    \textbf{SB Affinity Propagation} & 90.67($\pm$0.26) & 77.07($\pm$0.46) & 99.03($\pm$0.34) & 93.55($\pm$0.68) & 93.37($\pm$1.76) & 72.34($\pm$2.33) & \pmb{81.65($\pm$1.45)} & \pmb{87.92($\pm$1.24)} \\
    \noalign{\hrule height 1pt}
    \multicolumn{9}{c}{\textbf{Binary HDC using N-gram-based Encoding}} \\
    \noalign{\hrule height 1pt}
    \textbf{HDCluster} & 49.78($\pm$7.86) & 37.03($\pm$7.96) & 60.10($\pm$21.26) & 80.38($\pm$12.39) & 68.30($\pm$17.70) & 44.92($\pm$6.36) & 53.56($\pm$8.20) & 75.43($\pm$0.33) \\
    \textbf{SB Kmeans} & 56.88($\pm$3.77) & 54.16($\pm$2.20) & 68.77($\pm$20.36) & 92.38($\pm$2.60) & 85.99($\pm$0.14) & 55.43($\pm$2.18) & 64.96($\pm$2.30) & 75.38($\pm$0.00) \\
    \textbf{Bin Height} & 57.36($\pm$3.82) & 54.94($\pm$2.24) & 75.19($\pm$23.94) & 84.82($\pm$6.21) & 85.99($\pm$0.14) & 60.53($\pm$3.16) & 66.52($\pm$2.23) & 75.38($\pm$0.00) \\
    \textbf{Bin Width} & / & / & 7.70($\pm$18.84) & 92.71($\pm$0.61) & 86.00($\pm$0.11) & 55.12($\pm$1.97) & / & 75.41($\pm$0.26) \\
    \hline
    \specialcell[c]{\textbf{SB Kmeans (1-Pass)}} & 45.55($\pm$2.92) & 37.94($\pm$2.32) & 64.14($\pm$17.57) & 92.61($\pm$0.69) & 86.13($\pm$0.52) & 55.73($\pm$2.35) & 64.77($\pm$2.16) & 75.41($\pm$0.19) \\
     \specialcell[c]{\textbf{Bin Height (1-Pass)}} & 47.29($\pm$2.94) & 38.31($\pm$2.63) & 69.32($\pm$20.79) & 83.78($\pm$2.58) & 86.58($\pm$0.85) & 60.00($\pm$2.71) & 65.91($\pm$2.35) & 75.38($\pm$0.00) \\
    \specialcell[c]{\textbf{Bin Width (1-Pass)}}& / & / & 7.33($\pm$17.84) & 85.09($\pm$1.78) & 86.00($\pm$0.09) & 54.91($\pm$1.81) & / & 75.46($\pm$0.37) \\
    \hline
    \textbf{SB Affinity Propagation} & 90.68($\pm$1.01) & 76.65($\pm$0.59) & 89.22($\pm$16.18) & \pmb{94.62($\pm$0.41)} & \pmb{93.73($\pm$0.68)} & \pmb{72.70($\pm$2.01)} & 80.15($\pm$1.53) & 87.56($\pm$0.94) \\
    \noalign{\hrule height 1pt}
    \multicolumn{9}{c}{\textbf{Non-Binary HDC using Record-based Encoding}} \\
    \noalign{\hrule height 1pt}
    \textbf{HDCluster} & 50.26($\pm$8.28) & 38.97($\pm$6.41) & 58.32($\pm$15.28) & 80.15($\pm$13.89) & 68.68($\pm$16.27) & 44.12($\pm$6.41) & 55.46($\pm$9.69) & 75.38($\pm$0.00) \\
    \textbf{SB Kmeans} & 54.95($\pm$3.45) & 53.04($\pm$2.61) & 84.42($\pm$1.56) & 91.90($\pm$0.82) & 85.88($\pm$0.99) & 55.84($\pm$2.17) & 68.58($\pm$1.67) & 75.38($\pm$0.00) \\
    \textbf{Bin Height}  & 54.45($\pm$3.21) & 53.76($\pm$2.39) & 93.05($\pm$2.12) & 91.32($\pm$1.01) & 85.77($\pm$0.96) & 56.35($\pm$2.38) & 67.93($\pm$1.75) & 75.38($\pm$0.00) \\
    \textbf{Bin Width} & 55.51($\pm$2.76) & / & 77.77($\pm$7.12) & 92.32($\pm$0.58) & 86.83($\pm$1.42) & 55.15($\pm$0.78) & / & 75.44($\pm$0.31) \\
    \hline
    \specialcell[c]{\textbf{SB Kmeans (1-Pass)}} & 46.58($\pm$2.25) & 38.98($\pm$2.12) & 82.98($\pm$2.57) & 91.70($\pm$0.87) & 87.31($\pm$0.98) & 55.43($\pm$2.12) & 67.34($\pm$1.58) & 75.58($\pm$0.56) \\
     \specialcell[c]{\textbf{Bin Height  (1-Pass)}}& 49.22($\pm$2.01) & 39.82($\pm$2.22) & 89.04($\pm$3.89) & 90.28($\pm$1.32) & 86.85($\pm$1.03) & 57.36($\pm$2.09) & 67.38($\pm$1.78) & 75.38($\pm$0.00) \\
    \specialcell[c]{\textbf{Bin Width (1-Pass)}} & 39.17($\pm$2.44) & / & 66.35($\pm$6.50) & 92.22($\pm$0.53) & 87.95($\pm$1.04) & 54.72($\pm$1.46) & / & 76.70($\pm$1.23) \\
    \hline
    \textbf{SB Affinity Propagation} & 91.07($\pm$0.18) & 77.48($\pm$0.39) & 99.67($\pm$0.16) & 94.30($\pm$0.48) & 93.00($\pm$0.86) & 72.37($\pm$2.29) & 78.92($\pm$1.28) & 86.78($\pm$1.02) \\
    \noalign{\hrule height 1pt}
    \multicolumn{9}{c}{\textbf{Non-Binary HDC using N-gram-based Encoding}} \\
    \noalign{\hrule height 1pt}
    \textbf{HDCluster} & 50.16($\pm$9.00) & 38.92($\pm$7.35) & 59.86($\pm$17.24) & 79.99($\pm$13.95) & 67.74($\pm$16.83) & 44.40($\pm$6.41) & 54.66($\pm$10.14) & 75.39($\pm$0.02) \\
    \textbf{SB Kmeans} & 55.18($\pm$3.63) & 53.48($\pm$2.50) & 84.30($\pm$1.58) & 92.00($\pm$0.45) & 86.11($\pm$0.65) & 55.79($\pm$1.72) & 67.83($\pm$1.56) & 75.38($\pm$0.00) \\
    \textbf{Bin Height}  & 54.99($\pm$3.57) & 53.57($\pm$2.50) & 93.14($\pm$2.49) & 91.48($\pm$0.67) & 85.86($\pm$0.65) & 55.89($\pm$1.82) & 68.15($\pm$1.77) & 75.38($\pm$0.00) \\
    \textbf{Bin Width} & 55.46($\pm$3.27) & / & 74.86($\pm$6.50) & 92.39($\pm$0.26) & 86.99($\pm$1.13) & 55.00($\pm$0.47) & / & 75.46($\pm$0.38) \\
    \hline
    \specialcell[c]{\textbf{SB Kmeans (1-Pass)}} & 46.66($\pm$2.50) & 39.20($\pm$2.22) & 83.31($\pm$2.57) & 91.75($\pm$0.53) & 87.48($\pm$0.79) & 55.22($\pm$1.79) & 66.79($\pm$1.43) & 75.44($\pm$0.28) \\
     \specialcell[c]{\textbf{Bin Height  (1-Pass)}} & 48.82($\pm$2.64) & 39.67($\pm$2.54) & 89.66($\pm$3.52) & 90.42($\pm$1.09) & 86.83($\pm$0.73) & 56.91($\pm$1.72) & 67.54($\pm$1.87) & 75.38($\pm$0.00) \\
    \specialcell[c]{\textbf{Bin Width (1-Pass)}} & 41.27($\pm$2.92) & / & 67.90($\pm$5.46) & 92.39($\pm$0.23) & 87.80($\pm$0.77) & 54.77($\pm$0.98) & / & 76.53($\pm$1.10) \\
    \hline
    \textbf{SB Affinity Propagation} & \pmb{91.08($\pm$0.41)} & \pmb{77.53($\pm$0.45)} & \pmb{99.72($\pm$0.13)} & 94.33($\pm$0.35) & 93.11($\pm$0.79) & 72.30($\pm$2.52) & 79.15($\pm$1.27) & 86.46($\pm$0.42) \\
   \noalign{\hrule height 1pt}
   \multicolumn{9}{l}{\specialcell[b]{$^*$This table displays the clustering accuracy over 500 runs in the format of $[$mean ($\pm$ standard deviation)$]$. \\Symbol ``/'' indicates the result is not available.\\
   SB is short for ``similarity-based''.}}
    \end{tabular}\label{t:app:mean_std_ours_vs_hdcluster}
  \end{adjustbox}
\end{table*}

\section{Performance of Our Proposed HDC-based Clustering Algorithms.}
\begin{figure*}[ht]
\centering
\subfigure[Performance for clustering algorithms by binary HDC using N-gram-based encoding.]{
\includegraphics[width=0.95\linewidth]{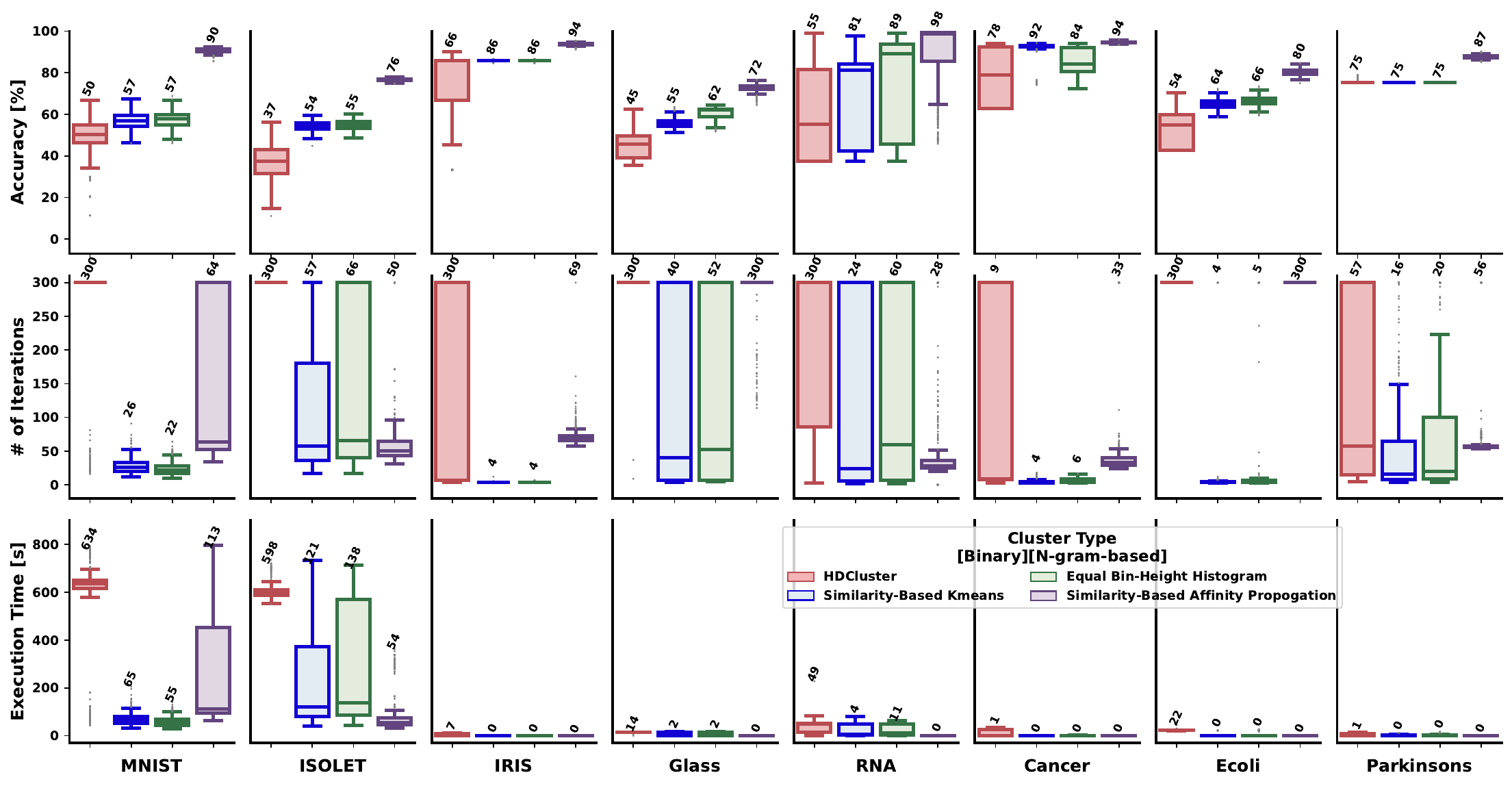}}\label{fig:per_vs_imani_ham_per}
\vspace{-8pt}
\subfigure[Performance for clustering algorithms by non-binary HDC using record-based encoding.]{
\includegraphics[width=0.95\linewidth]{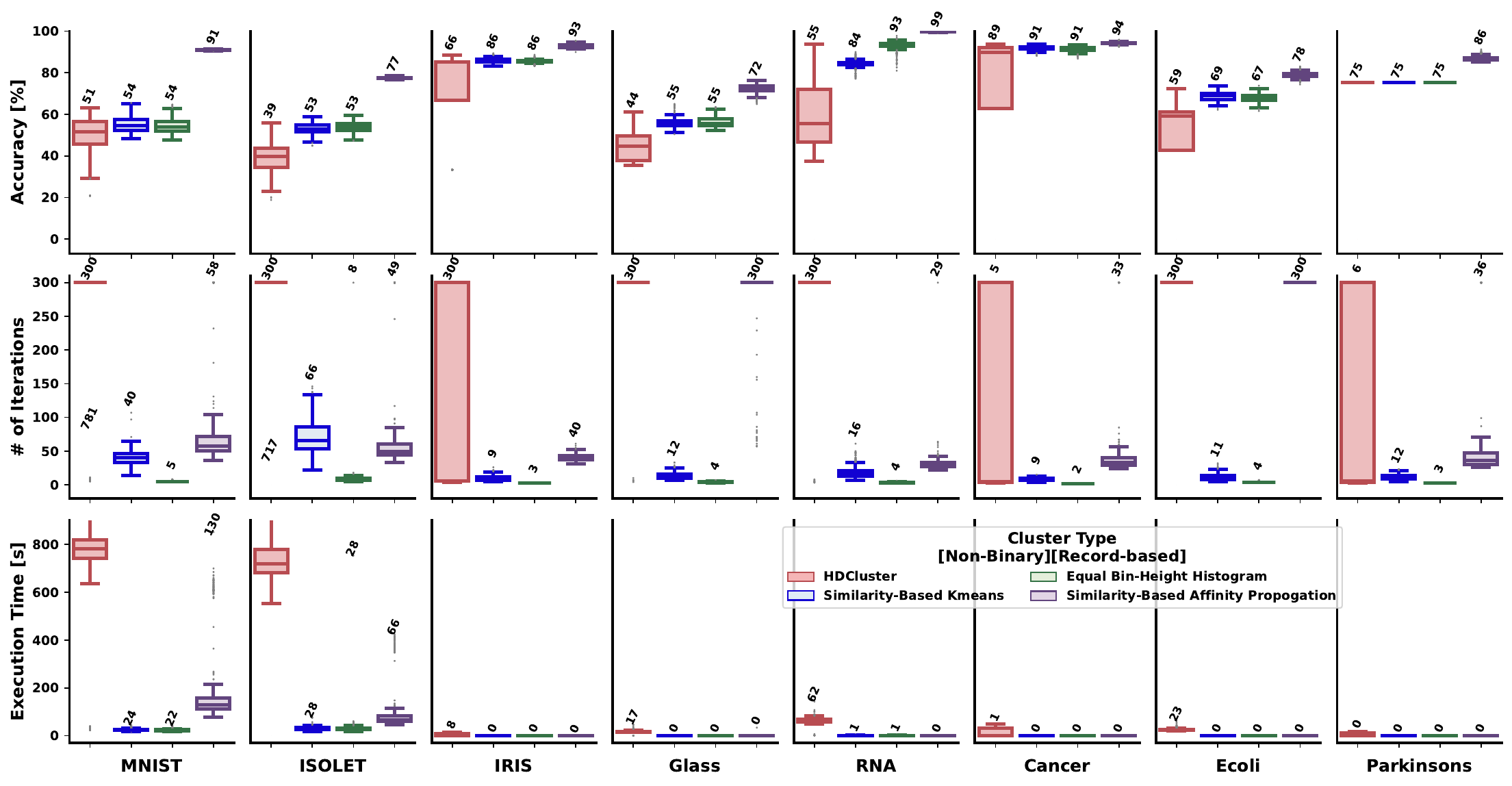}}\label{fig:per_vs_imani_cos_re}
\caption{Comparison of proposed algorithms with HDCluster (Continued from main content). For boxplots, the median values over 500 runs are annotated on the top.}\label{fig:performace_vs_imani(conti.)}
\vspace{-8pt}
\end{figure*}

Figure \ref{fig:performace_vs_imani(conti.)} shows the performance of our proposed algorithms in the other three cases. 

\setcounter{figure}{9}
\begin{figure*}[ht]
\setcounter{subfigure}{2}
\centering
\subfigure[Performance for clustering algorithms by non-binary HDC using N-gram-based encoding.]{
\includegraphics[width=0.95\linewidth]{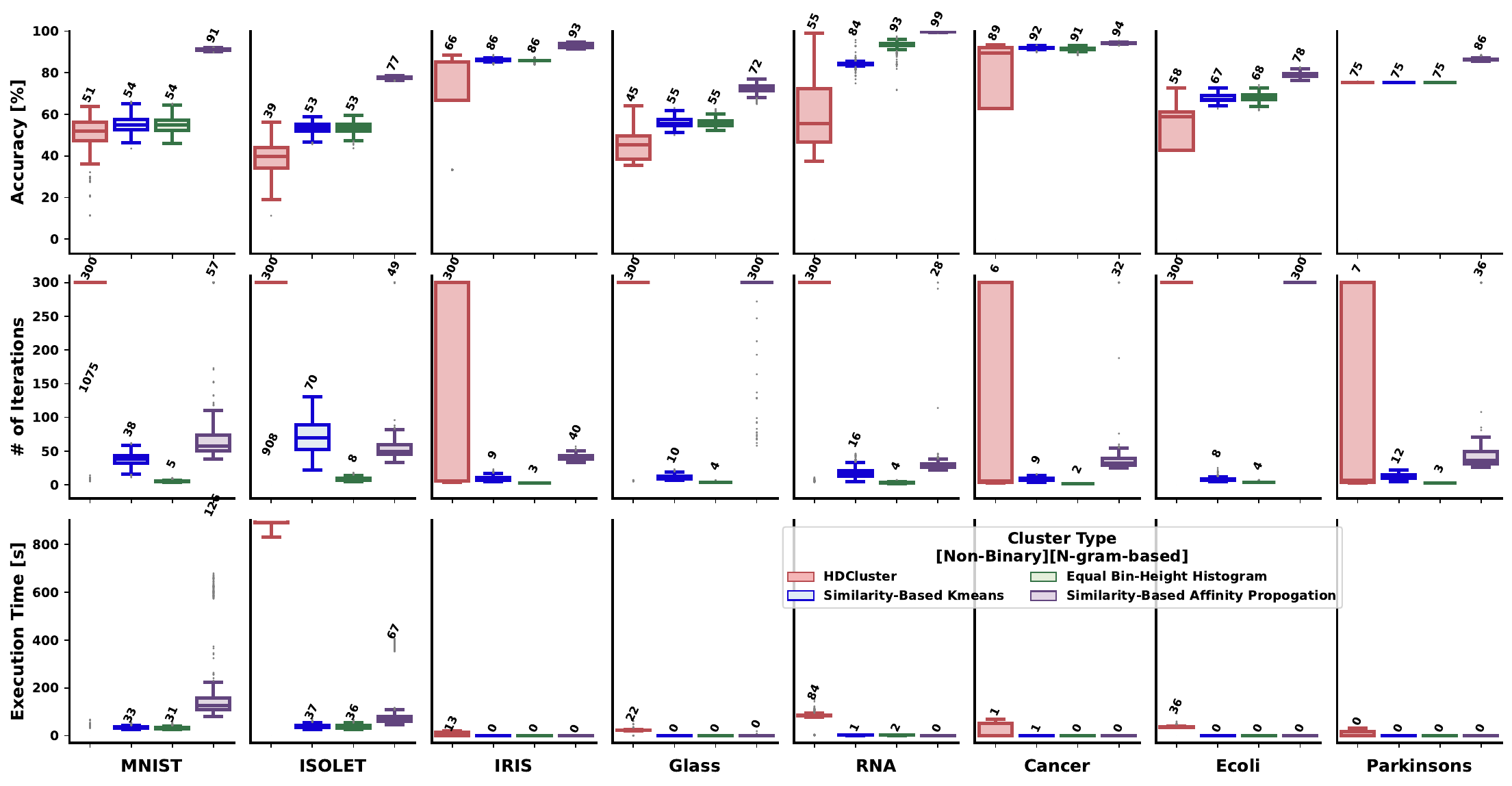}}\label{fig:per_vs_imani_cos_per}
\caption{Comparison of proposed algorithms with HDCluster (Continued). For boxplots, the median values over 500 runs are annotated on the top. }
\vspace{-8pt}
\end{figure*}


\section{Performance of One-pass Clustering using Our Proposed HDC-based Clustering Algorithms.}
\begin{figure*}[t]
\centering
\subfigure[Binary HDC using N-gram-based encoding.]{
\includegraphics[width=0.9\linewidth]{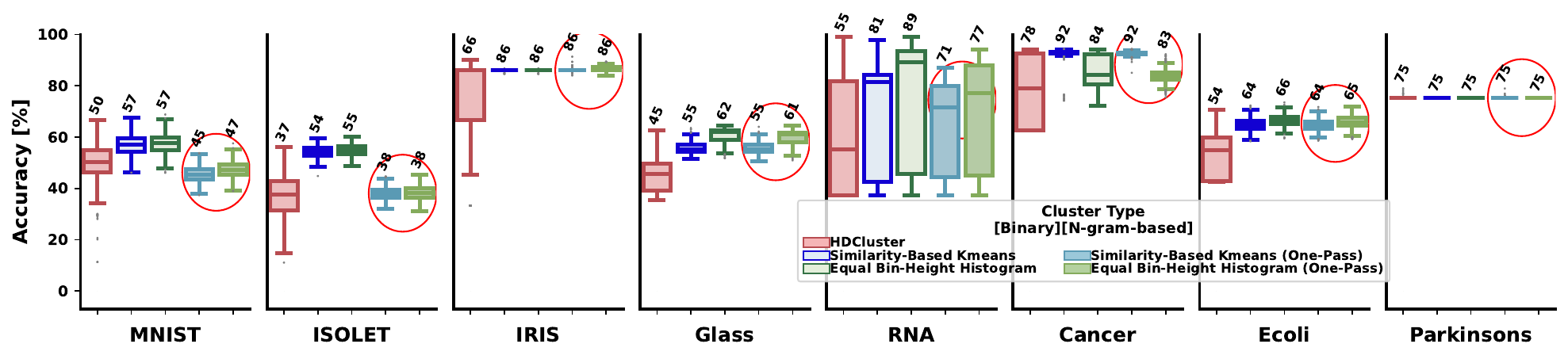}}\label{fig:1_shot_ham_per}
\vspace{-10pt}
\subfigure[Non-binary HDC using record-based encoding.]{
\includegraphics[width=0.9\linewidth]{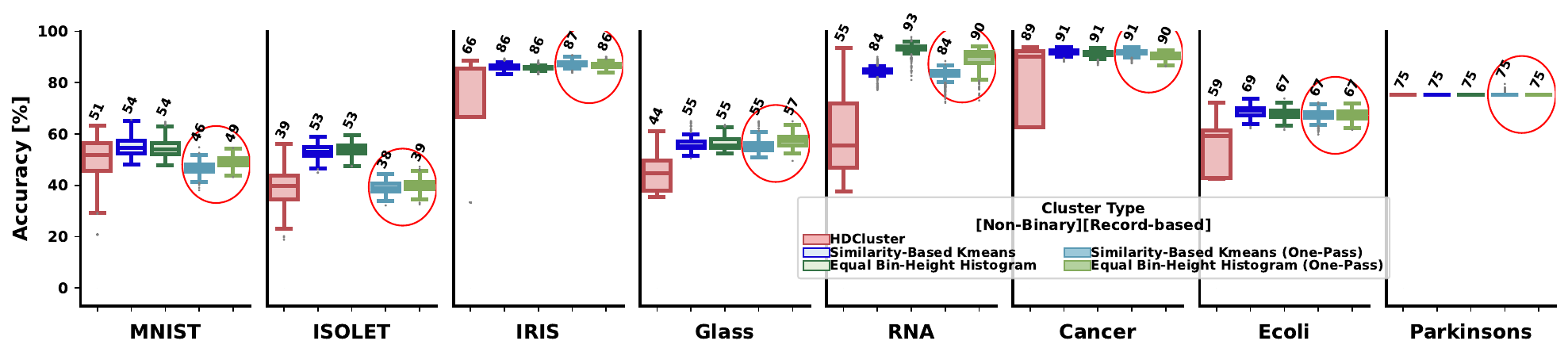}}\label{fig:1_shot_cos_re}
\vspace{-12pt}
\subfigure[Non-binary HDC using N-gram-based encoding.]{
\includegraphics[width=0.9\linewidth]{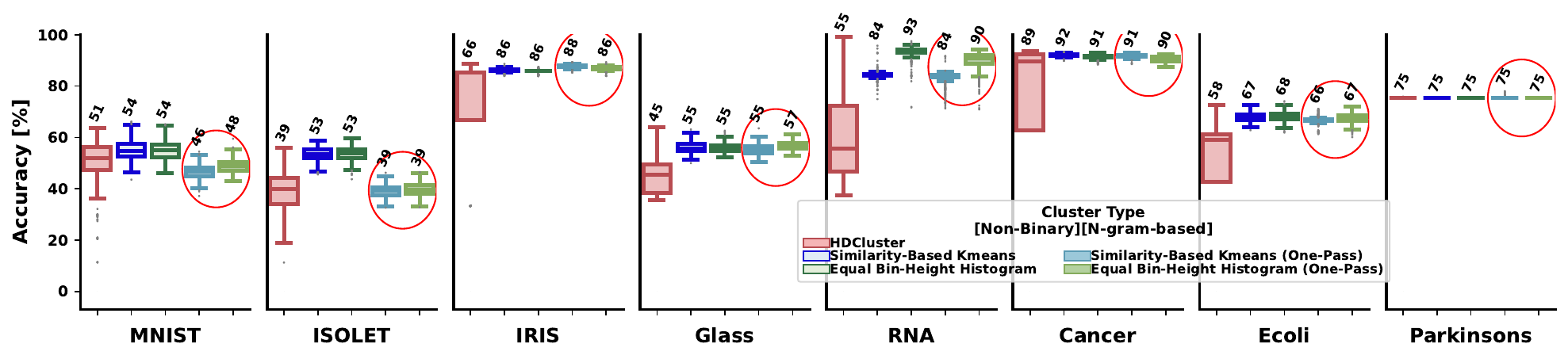}}\label{fig:1_shot_cos_per}
\caption{Comparison of one-pass clustering with both the updated clusters and HDCluster (Continued from main content). For boxplots, the median values over 500 runs are annotated on the top.}\label{fig:performance_1shot(cont.)}
\vspace{-8pt}
\end{figure*}

Figure \ref{fig:performance_1shot(cont.)} shows the other three cases of the one-pass clustering performance using our algorithms as compared to HDCluster.

\section{Performance of Original Data vs. Encoded Data for Traditional Clustering Algorithms.}\label{app:ori_vs_query}

\begin{figure*}[ht]
\centering
\subfigure[Binary HDC using N-gram-based encoding.]{
\includegraphics[width=0.98\linewidth]{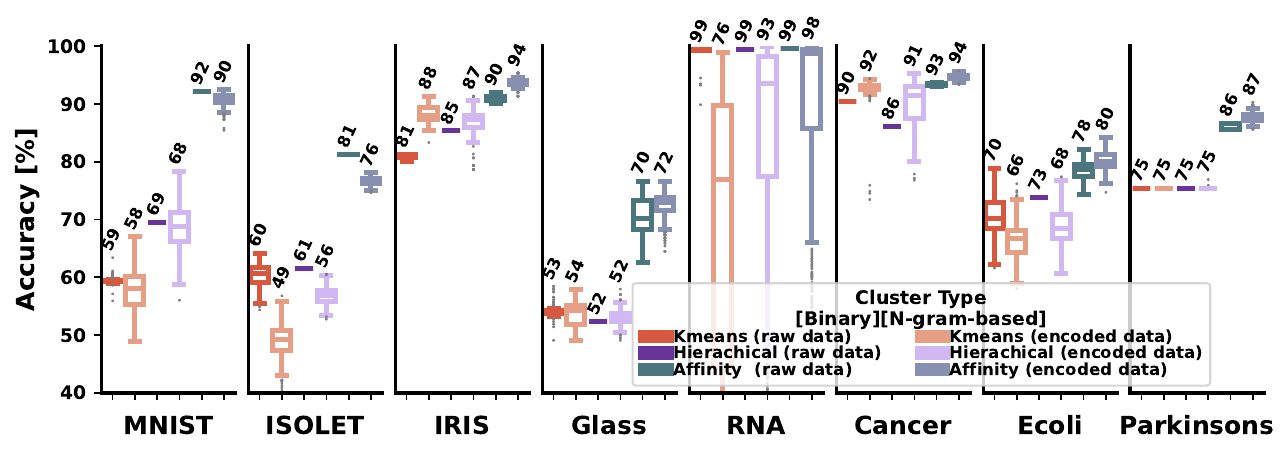}}\label{fig:query_ham_per}
\subfigure[Non-binary HDC using record-based encoding.]{
\includegraphics[width=0.98\linewidth]{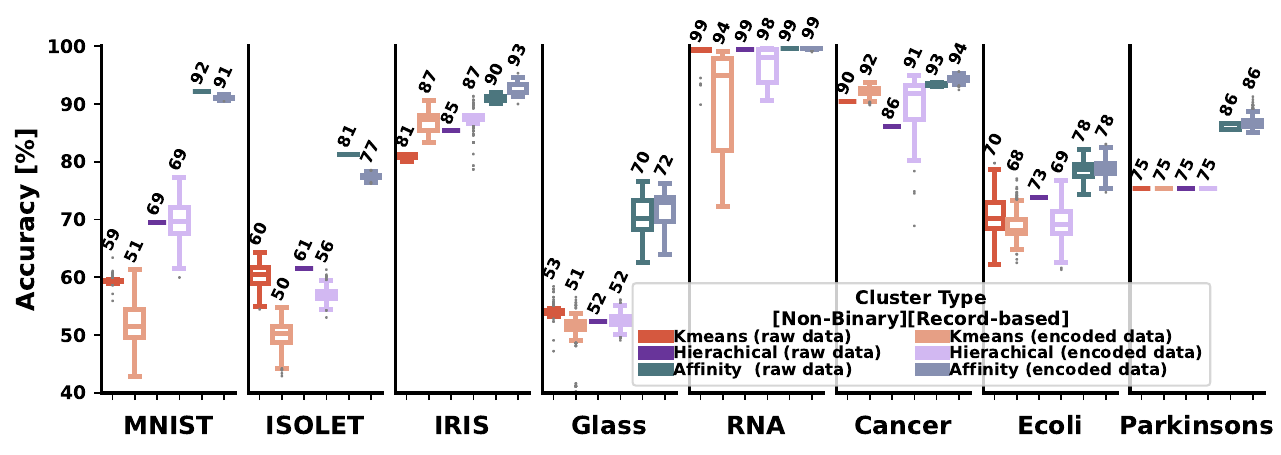}}\label{fig:query_cos_re}
\subfigure[Non-binary HDC using N-gram-based encoding.]{
\includegraphics[width=0.98\linewidth]{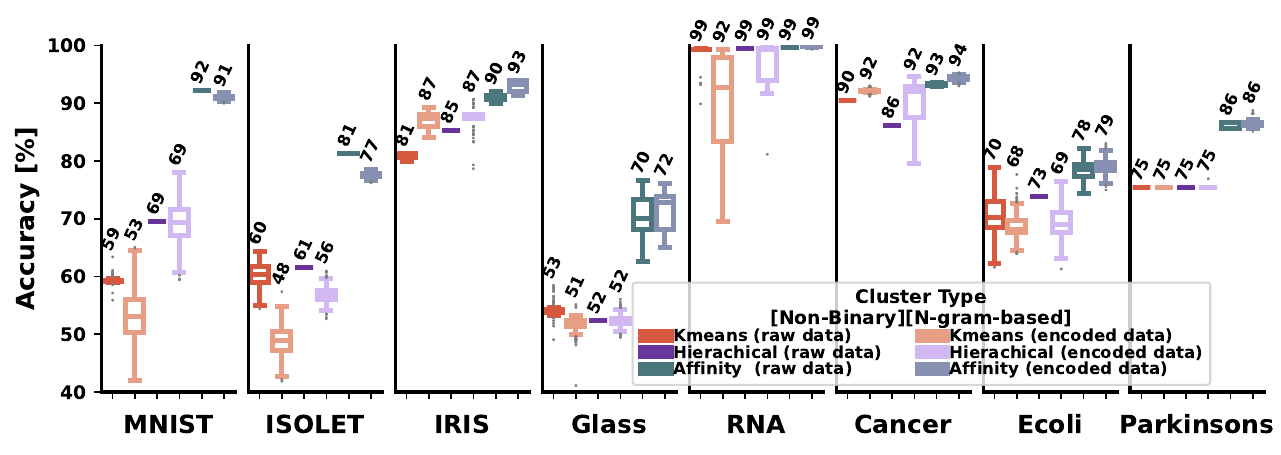}}\label{fig:query_cos_per}
\caption{Comparison of clustering using original data and encoded data (Continued from main content). For boxplots, the median values over 500 runs are annotated on the top.}\label{fig:performance_vs_query(conti.)}
\vspace{-8pt}
\end{figure*}

Figure \ref{fig:performance_vs_query(conti.)} shows the other three cases' results of the clustering performance for a comparison of original data with the encoded data.

\clearpage
\balance
\section*{Acknowledgment}
 The authors thank Xingyi Liu and Sai Sanjeet for numerous useful discussions. 

\ifCLASSOPTIONcaptionsoff
  \newpage
\fi


\footnotesize
\bibliographystyle{IEEEtran}
\bibliography{IEEEabrv,mybib}
\flushend

\begin{IEEEbiography}[{\includegraphics[width=1in,height=1.25in,clip,keepaspectratio]{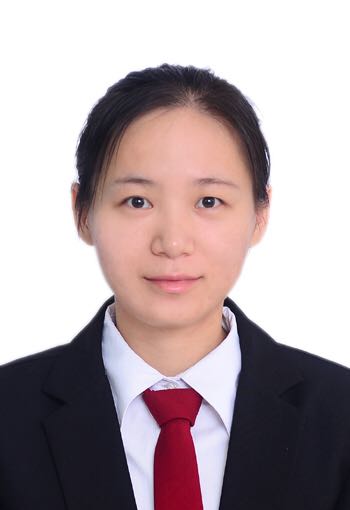}}]{Lulu Ge}
received a B.S. degree from Nanjing University of Posts and Telecommunications (NJUPT), Nanjing, China, in 2015, and an M.S. degree from the Southeast University, Nanjing, China, in 2018. She is currently pursuing her Ph.D. degree in electrical engineering at the University of Minnesota, Minneapolis, MN, USA. Her research interests include hyperdimensional computing and machine learning.
\end{IEEEbiography}

\begin{IEEEbiography}[{\includegraphics[width=1in,height=1.25in,clip,keepaspectratio]{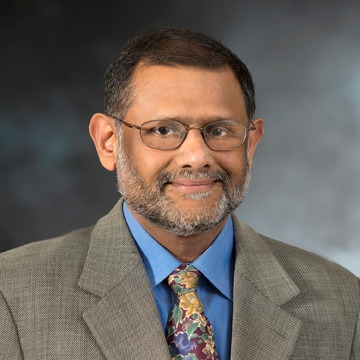}}]{Keshab K. Parhi} (Fellow, IEEE)
received the B.Tech. degree from the Indian Institute
of Technology (IIT), Kharagpur, in 1982, the M.S.E.E. degree from the
University of Pennsylvania, Philadelphia, in 1984, and the Ph.D.
degree from the University of California, Berkeley, in 1988. He has
been with the University of Minnesota, Minneapolis, since 1988, where
he is currently Erwin A. Kelen Chair and Distinguished McKnight University Professor
in the Department of
Electrical and Computer Engineering. He has published over 700 papers,
is the inventor of 34 patents, and has authored the textbook VLSI
Digital Signal Processing Systems (Wiley, 1999). His current research
addresses VLSI architecture design of machine learning and signal processing systems,
hardware security, and data-driven neuroengineering and neuroscience.
Dr. Parhi is the recipient of numerous awards including the
2017 Mac Van Valkenburg Award and the 2012 Charles A. Desoer Technical
Achievement award from the IEEE Circuits and Systems Society, 2003
IEEE Kiyo Tomiyasu Technical Field Award, and a Golden Jubilee medal
from the IEEE Circuits and Systems Society in 2000. He served as the
Editor-in-Chief of the IEEE Trans. Circuits and Systems, Part-I during
2004 and 2005. He is a
Fellow of the American Association for the Advancement of Science
(AAAS), the Association for Computing Machinery (ACM), the American Institute of
Medical and Biological Engineering (AIMBE), and the National Academy of Inventors (NAI).
\end{IEEEbiography}

\end{document}